\RequirePackage{fix-cm}
\documentclass[smallextended,envcountsect,final=]{svjour3} 
\smartqed 
\usepackage{algorithm}
\usepackage{algorithmic}
\usepackage{chemfig}
\usepackage{bm}
\usepackage{subfigure}
\usepackage{booktabs}
\usepackage{multirow}
\usepackage{url}
\usepackage{graphicx} 
\usepackage{tikz}
\usepackage{setspace}
\usepackage{titlesec}
\usepackage{pgfplots}
\usepgfplotslibrary{groupplots}
\pgfplotsset{width=0.8\columnwidth, height=0.49\columnwidth,compat=1.9}
\usepackage{amsmath}
\allowdisplaybreaks
\newcommand\mtlarge{\fontsize{8pt}{10pt}\selectfont}
\newcommand\markSize{1.5}

\pgfplotsset{every axis/.append style={
xlabel={$x$},          
ylabel={$y$},          
label style={font=\mtlarge},
tick label style={font=\mtlarge},
legend style={font=\mtlarge}
}}

\titleformat{\subsection}{\bfseries}{\thesubsection}{0.4em}{}
\usepackage{enumitem}
\setitemize[1]{itemsep=2.5pt,partopsep=0.5pt,parsep=\parskip,topsep=2.5pt}
\newcommand{\IB}{\mathcal{I}_\mathcal{B}}
\newcommand{\IW}{\mathcal{I}_\mathcal{W}}

\usepackage{amsmath,amssymb,graphicx,mathtools}
\newcommand{\defeq}{\vcentcolon=}

\usepackage{xcolor}
\usepackage{amssymb}
\usepackage{pifont}
\newcommand{\xmark}{\ding{55}}

\newcommand{\citer}{k}
\newcommand{\tcon}{K}

\newcommand{\be}{\begin{equation}}
\newcommand{\ee}{\end{equation}}
\newcommand{\bee}{\begin{eqnarray}}
\newcommand{\eee}{\end{eqnarray}}
\newcommand{\bse}{\begin{subequations}}
\newcommand{\ese}{\end{subequations}}

\newcommand{\Hil}{\mathcal{H}}

\newcommand{\betac}{\beta}

\journalname{Noname}

\newcommand{\cnj}[1]{\textcolor{red}{}} 

\newcommand{\wxu}[1]{\textcolor{red}{}} 

\newtheorem{assumption}[theorem]{Assumption}
\spnewtheorem{obs}{Observation}[section]{\bf}{\it}
\spnewtheorem{assump}{Assumption}[section]{\bf}{\it}
\spnewtheorem{conject}{Conjecture}[section]{\bf}{\it}
\spnewtheorem{myExample}{Example}[section]{\bf}{\it}
\spnewtheorem{myRemark}{Remark}[section]{\bf}{\it}

\ifodd 1
    
    \newcommand{\com}[1]{\textbf{\color{red} (COMMENT: #1)}} 
\else
    
    \newcommand{\com}[1]{}
\fi

\title{Bayesian Optimization of Expensive Nested Grey-Box Functions}
\author{Wenjie Xu \and Yuning Jiang \and \\Bratislav Svetozarevic\and Colin N. Jones}

\institute{
W. Xu, Y. Jiang, C. Jones \at
Automatic Control Laboratory\\
EPFL, Switzerland\\
\email{wenjie.xu, yuning.jiang, colin.jones@epfl.ch}\\[0.2cm]
W. Xu and B. Svetozarevic are also \\
with Swiss Federal Laboratories\\
for Materials Science and Technology (Empa)\\
\email{bratislav.svetozarevic@empa.ch}
}

\begin{document}

\maketitle

\begin{abstract}
We consider the problem of optimizing a \emph{grey-box} objective function, i.e., nested function composed of both \emph{black-box} and \emph{white-box} functions. A general formulation for such grey-box problems is given, which covers the existing grey-box optimization formulations as special cases. We then design an \emph{optimism}-driven algorithm to solve it. Under certain regularity assumptions, our algorithm achieves similar regret bound as that for the standard black-box Bayesian optimization algorithm, up to a constant multiplicative term depending on the Lipschitz constants of the functions considered. We further extend our method to the constrained case and discuss special cases. For the commonly used kernel functions, the regret bounds allow us to derive a convergence rate to the optimal solution. Experimental results show that our grey-box optimization method empirically improves the speed of finding the global optimal solution significantly, as compared to the standard black-box optimization algorithm.    
\end{abstract}

\section{Introduction}
We consider the problem of optimizing a \emph{grey-box} objective function, i.e., nested function composed of both \emph{black-box} and \emph{white-box} functions. Such problems arise from a wide variety of applications. For example, while we optimize an economic objective subject to some operational constraints in process control~\cite{del2021real,xu2022config}, the objective can be a bilinear function in known price variable and product fraction variables, where the latter are unknown black-box functions of the tunable operational parameters.

To solve such problems, one may ignore the grey-box structure and treat the nested function as a pure input-to-output black-box function. Bayesian optimization~(BO)~\cite{frazier2018tutorial} has been shown to be an effective method to solve the purely black-box optimization problem~\cite{snoek2012practical}. However, the agnostic application of the black-box BO method to the grey-box problem has several limitations, 
\begin{itemize}
    \item Fails to utilize the intermediate observations and the known structure of how different \emph{black-box} and \emph{white-box} elements influence each other. However, it has been shown that such structural information can empirically boost the performance of Bayesian optimization~\cite{astudillo2019bayesian}.  
    
    \item May violate the regularity assumptions necessary for convergence guarantee. For example, a common assumption for convergence guarantee of the black-box Bayesian optimization method is the bounded norm inside a reproducing kernel Hilbert space~\cite{srinivas2012information,xu2022config}. This assumption can be violated for the grey-box function with a discontinuous white-box element.      
\end{itemize}

In this paper, we consider a general grey-box optimization problem and propose an \emph{optimism}-driven algorithm to solve it. Specifically, our contributions are summarized as follows:
\begin{itemize}
    \item A general formulation for grey-box optimization problems is given, which covers the existing grey-box optimization formulations~\cite{astudillo2019bayesian,astudillo2021bayesian,lu2023no} as special cases. 
    
    \item An \emph{optimism}-driven algorithm is designed to solve it. Under certain regularity assumptions, our algorithm achieves similar regret bound as that for the standard black-box Bayesian optimization algorithm, up to a constant multiplicative term depending on the Lipschitz constants of the functions considered. For the commonly used kernel functions, the regret bounds allow us to derive a convergence rate to the optimal solution. 
    
    \item Our method is further extended to the constrained case and several special cases are discussed. Experimental results show that our grey-box optimization method empirically improves the speed of finding the global optimal solution significantly, as compared to the standard black-box optimization algorithm.    
\end{itemize}

\section{Related work}
The problem we consider in this paper follows the research line of Bayesian optimization~(BO), a framework of efficient global optimization of expensive-to-evaluate black-box function~\cite{zhilinskas1975single,movckus1975bayesian,jones1998efficient,xu2022lower}. Standard Bayesian optimization~\cite{frazier2018tutorial} assumes that the expensive-to-evaluate functions are purely black-box. That is to say, the function is not known explicitly but only provides an oracle that can be queried to get the zero-order function evaluation. There are also works~\cite{gardner2014bayesian,gelbart2014bayesian,xu2021vabo,xu2022constrained,xu2023violation} that extend the black-box Bayesian optimization to the constrained setting. 

More recently, grey-box Bayesian optimization has been receiving more and more attention. For example, \cite{kandasamy2017multi} and \cite{wu2020practical} use cheap approximations of the black-box functions to do multi-fidelity BO. \cite{toscano2018bayesian,cakmak2020bayesian} propose a Bayesian optimization method for the unknown black-box function with integration structure. More generally, \cite{astudillo2019bayesian,lu2023no} considers Bayesian optimization of a composite function, with an expensive-to-evaluate inner black-box function and cheap-to-evaluate outer known function. \cite{astudillo2019bayesian} further considers Bayesian optimization of a network of functions.  

Another line of research considers causal Bayesian optimization~\cite{aglietti2020causal,aglietti2021dynamic,aglietti2023constrained,sussex2022model}, where the relationship between intervention/input variables and target/output variables are modeled as a causal graph. However, this line of research results only considered the probabilistic modelling of the variables, but did not consider the known intermediate white-box functions.    

Existing works on grey-box Bayesian optimization either only consider special structures such as addition, integration~\cite{toscano2018bayesian,cakmak2020bayesian}, or one known function composed with black-box functions~\cite{astudillo2019bayesian,lu2023no}, or can only provide very weak theoretical guarantee~(such as the consistency of finding the global optimum). Our work can be regarded as a general extension of the research line for grey-box Bayesian optimization. Furthermore, we design an \emph{optimism}-driven algorithm to solve the problem with a global convergence rate guarantee. To highlight the contributions of our work, we compare our method with the other state-of-the-art methods in Tab.~\ref{tab:comparison}. 

\begin{table*}[htbp!]
\renewcommand{\arraystretch}{1.2}
\centering

{\caption{The comparison of our method to existing state-of-the-art grey-box Bayesian optimization methods, where `hybrid' means the grey-box function is composed of both unknown black-box functions and known white-box functions.} 
\label{tab:comparison}
\resizebox{0.99\columnwidth}{!}{
\begin{tabular}{|c|c|c|c|c|}
\hline 
\textbf{Works} & \begin{tabular}{@{}c@{}}\textbf{Regret}\\ \textbf{Bound}\end{tabular}&\begin{tabular}{@{}c@{}}\textbf{Constraints}\\ \textbf{Consideration}\end{tabular} & \begin{tabular}{@{}c@{}}\textbf{Multilevel}\\{\textbf{Modelling}}\end{tabular} & \begin{tabular}{@{}c@{}}\textbf{Function}\\\textbf{Type}\end{tabular}\\
\hline 
\begin{tabular}{@{}c@{}}
\textsf{Composite EI}~
\cite{astudillo2019bayesian}, etc.
\end{tabular}& \xmark & \xmark & \xmark& hybrid \\
\hline 
\begin{tabular}{@{}c@{}}
\textsf{BO of function network}~
\cite{astudillo2021bayesian}, etc.
\end{tabular}& \xmark & \xmark & \checkmark& hybrid \\
\hline 
\begin{tabular}{@{}c@{}}
\textsf{CUQB}~
\cite{lu2023no}
\end{tabular}& \checkmark & \checkmark & \xmark& hybrid \\
\hline 
\begin{tabular}{@{}c@{}}
\textsf{Causal BO}~
\cite{aglietti2020causal,aglietti2021dynamic}
\end{tabular}& \xmark& \xmark & \checkmark & black-box \\
\hline
\begin{tabular}{@{}c@{}}
\textsf{Constrained Causal BO}~
\cite{aglietti2023constrained}
\end{tabular}& \xmark& \checkmark & \checkmark & black-box \\
\hline
\begin{tabular}{@{}c@{}}
\textsf{Model-based Causal BO}~
\cite{sussex2022model}
\end{tabular}& \checkmark& \xmark & \checkmark & black-box \\
\hline
\begin{tabular}{@{}c@{}}
\textsf{Ours}~
\end{tabular} & \checkmark& \checkmark & \checkmark  & hybrid \\
\hline 
\end{tabular}
} 
}
\end{table*}
 
\section{Grey-box modelling}
We use a hybrid factorable function to model the grey-box function. Sec.~\ref{subsec:alg_repr} will give the algebraic representation of the grey-box function, and Sec.~\ref{subsec:graph_repr} will give its corresponding graph representation. Before introducing our grey-box modelling approach, we first clarify what we will mean by \emph{black-box function} and \emph{white-box} function. An expensive unknown function to which we can only query zero-order function evaluation is referred to as a \emph{black-box} function. A cheap known function which has explicit mathematical expression is referred to as a \emph{white-box} function.  

\subsection{Algebraic representation}
\label{subsec:alg_repr}
We first introduce the algebraic representation of the grey-box function $f:\mathbb{R}^n\to\mathbb{R}$. The modelling idea is introducing a set of intermediate variables $z_i,i\in[m]$ and intermediate functions $\phi_{i-1},i\in[m]$. Formally, the following definitions are needed.
\begin{definition}[Augmented State]
   $\forall x\in\mathbb{R}^n, z\in\mathbb{R}^m$, we define the augmented state by,
   $$
   s_0=x=\left[\begin{array}{c}
x_1 \\
\vdots \\
x_n
\end{array}\right], s_1=\left[\begin{array}{c}
x_1 \\
\vdots \\
x_{n}\\
z_1
\end{array}\right], \ldots ., s_m=\left[\begin{array}{c}
x_1 \\
\vdots \\
x_{n} \\
z_1\\
\vdots \\
z_m
\end{array}\right].
   $$
   \label{def:aug_state}
\end{definition}

\begin{definition}[Augmented Elementary Function]
    Augmented elementary function $\Phi_i: \mathbb{R}^{n+i} \rightarrow \mathbb{R}^{n+i+1}$ is given as
    \[
    \Phi_i\left(s_i\right)=\left[\begin{array}{c}
    x_1 \\
    \vdots \\
    x_{n} \\
    z_{1} \\
    \vdots \\
    z_{i}\\
    \phi_i\left(x_1,\cdots, x_{n}, z_1,\cdots, z_i\right)
    \end{array}\right], s_{i+1}=\Phi_i\left(s_i\right),
    \]
    where $\phi_i:\mathbb{R}^{n+i} \rightarrow \mathbb{R}$ is either a known white-box function or an unknown black-box function. Thus, it can be seen that $z_{i+1}=\phi_i(s_i),i\in\{0,\cdots,m-1\}$.
    \label{def:aug_ele_func}
\end{definition}
\begin{myRemark}
   Note that here $\phi_i$ takes all the variables in $\{x_1,x_2,\cdots,x_n,z_1,\cdots,z_m\}$ as input. In practice, $\phi_i$ can be a function of a subset of variables $\{x_1,\cdots, x_{n}, z_1,\cdots, z_i\}$. All the following algorithms and analyses can be easily adapted to the practical setting of the subset input variables. For the simplicity of notation, we will focus on the setting that $\phi_i$ takes the whole $s_i$ as the input in the description of the algorithm and the theoretical analysis.    
\end{myRemark}
We can now define the \emph{grey-box} objective function by
\begin{equation}
   f(x)=e_{n+m}^T\Phi_{m-1}\left(\Phi_{m-2}\left(\cdots \Phi_1\left(\Phi_0(x)\right)\right)\right), \label{eq:grey_box_func} 
\end{equation}
where $e_{n+m}\in\mathbb{R}^{n+m}$ is the vector with the last element being $1$ and all the other elements being $0$. We use $\mathcal{I}_\mathcal{W}\subset\{0,\cdots,m-1\}$~($\mathcal{I}_\mathcal{B}\subset\{0,\cdots,m-1\}$, resp.) to denote the index set of the known white-box (unknown black-box, resp.) functions among $\phi_i,i\in\{0,\cdots,m-1\}$. It holds that $\mathcal{I}_\mathcal{W}\cup\mathcal{I}_\mathcal{B}=\{0,\cdots,m-1\}$.   

\subsection{Graph representation}
\label{subsec:graph_repr}
\begin{figure}[htbp]
    \centering
    \includegraphics[width=0.6\textwidth]{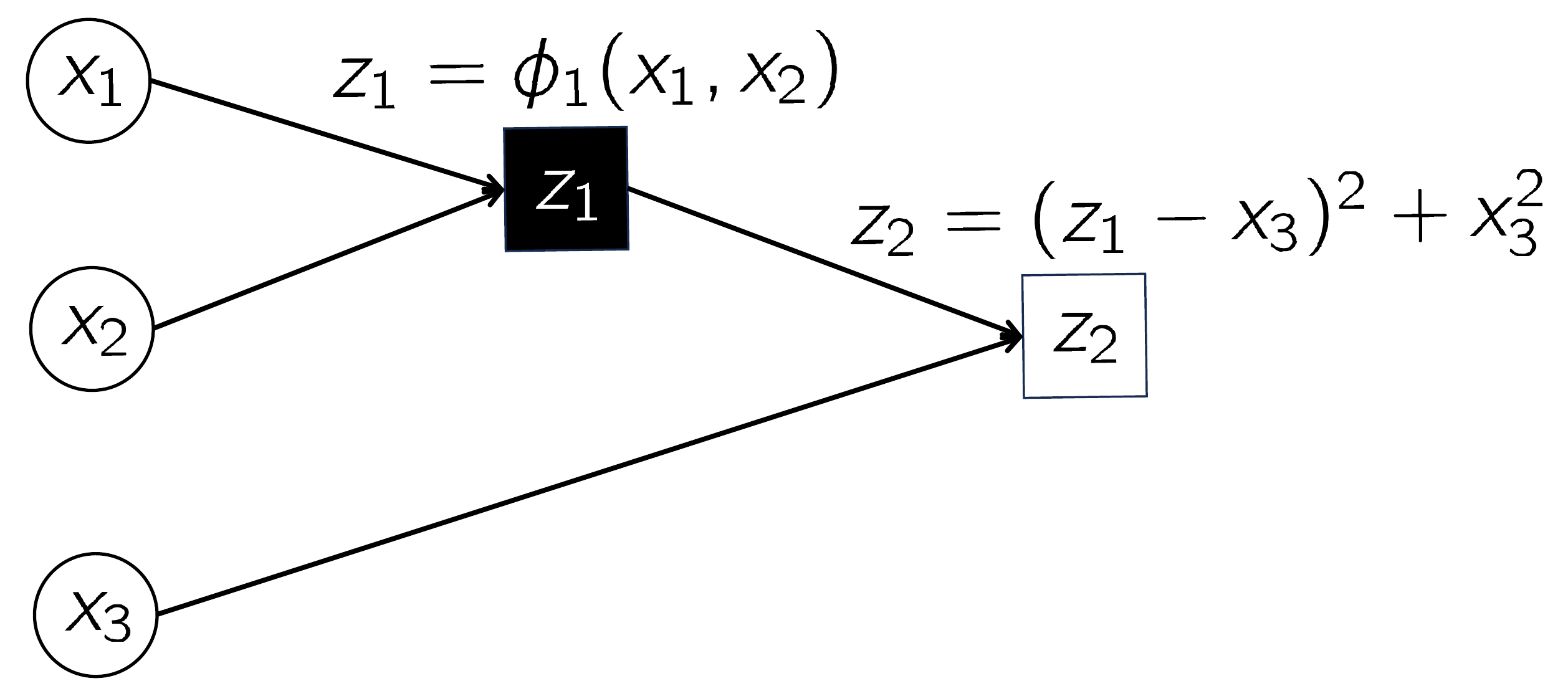}
    \caption{A simple example of the graph representation of a nested grey-box function.}
    \label{fig:grey_box_repre}
\end{figure}

The nested grey-box function can also be represented as an acyclic-directed graph $\mathcal{G}\coloneqq(\mathcal{V}, \mathcal{E})$. $\mathcal{V}=\mathcal{V}_x\cup\mathcal{V}_\phi$ is the vertices set, where $\mathcal{V}_x$ corresponds to the input variables $(x_i)_{i=1}^n$ and $\mathcal{V}_\phi$ corresponds to the intermediate variables $(z_i)_{i=1}^m$ that are outputs of the functions $(\phi_i)_{i=0}^{m-1}$. $(v_i,v_j)\in\mathcal{E}$ if and only if the variable of $v_j$ is either a black-box or white-box function that takes the variable of $v_i$ as one of the inputs. For example, Fig.~\ref{fig:grey_box_repre} gives the graph representation of a simple nested grey-box function $f(x)=(\phi_1(x_1, x_2)-x_3)^2+x_3^2$, where $\phi_1$ is a black-box function. 

\subsection{Some examples}
We now give examples to demonstrate that our grey-box modeling captures a rich class of functions. 
\begin{enumerate}
    \item {Additive black-box functions with different input variables.}
\begin{equation}
   f(x) = \phi_1(x_1, x_2)+\phi_2(x_2, x_3)+\phi_3(x_1, x_3),  
\end{equation}
where $\phi_i,i\in[3]$ are all black-box functions.

\item {Black-box functions composed with white box functions.}
\begin{equation}
    f(x) = \phi_1(x)^2+\phi_2(x)^2, 
\end{equation}
where $\phi_1$ and $\phi_2$ are black-box functions.

\item {Hybrid chain of black-box/white-box functions}
\begin{equation}
    f(x) = \phi_3\left(\phi_1(x)^2+3\phi_1(x)-3\right),
\end{equation}
where $\phi_1$ and $\phi_3$ are black-box functions.
\end{enumerate}

\section{Problem formulation}
We consider the following unconstrained global optimization problem of grey-box functions. 
\begin{subequations}
\label{eqn:prob_form}
\begin{eqnarray}
\min_{x\in\mathcal{X}}\quad &&f(x),\label{eqn:constrained_form}\\
\textrm{subject to}&& g_\citer(x)\leq 0, \citer\in[\tcon],
\end{eqnarray}
\end{subequations}
where $\mathcal{X}\subset\mathbb{R}^n$ is a known compact subset of $\mathbb{R}^n$, $f:\mathbb{R}^n\to\mathbb{R}$ is a grey-box function as given in Eq.~\eqref{eq:grey_box_func}, $g_k:\mathbb{R}^n\to\mathbb{R}$ is a constraint grey-box function similarly defined and $K$ is the number of constraints.

   Eq.~\eqref{eqn:prob_form} is a generic formulation of constrained grey-box optimization problem. This section focuses on the unconstrained case, where $K=0$ and we focus on $\min_{x\in\mathcal{X}} f(x)$. We will show later in Sec.~\ref{sec:ext_cons} that our method can easily be extended to the constrained case by adopting a similar algorithmic idea as proposed in~\cite{xu2022constrained}.   

At step $t$, with the sample point $x^t$, there is a vector of augmented states $\{z^t_{i+1}\}_{i\in\mathcal{I}_\mathcal{B}}$, where $z^t_{i+1}$ is the output of the black-box function $\phi_i$ with input $s^{t}_{i}$. The relationship of $(x^t, z^t, s^t)$ is as given in Def.~\ref{def:aug_state} and Def.~\ref{def:aug_ele_func}. Note that $z_{i+1}^t$ may not be observable, or the observation is corrupted by noise in practice. 

Similar to~\cite{xu2022constrained}, we make some regularity assumptions on the grey-box optimization problem~\eqref{eqn:prob_form}.
\begin{assumption}
Set $\mathcal{X}$ is compact. 
\label{assump:support_set}
\end{assumption}
\begin{assumption}
\label{assump:bounded_norm}
Functions $\phi_i\in\Hil_{k_i}, i\in\mathcal{I}_\mathcal{B}$, where $k_i:\mathbb{R}^{n_i}\times\mathbb{R}^{n_i}\to \mathbb{R}, i\in\IB$ are symmetric, positive-semidefinite kernel functions, $\Hil_{k_i}$ are the corresponding reproducing kernel Hilbert spaces~(RKHSs, see~\cite{scholkopf2001generalized}), and $n_i$ is the dimension of the input variables for the function $\phi_i$. Furthermore, we assume $\|\phi_i\|_{k_i}\leq B_i, i\in\IB$, where $\|\cdot\|_{k_i}$ is the norm induced by the inner product in the corresponding reproducing kernel Hilbert space.  
\end{assumption}
\begin{assumption}
\label{assump:regular_kernel}
 $\forall s_i\in\mathbb{R}^{n+i},\,i\in\IB$, $k_i(s_i, s_i)\leq 1$.
\end{assumption}

\begin{assumption}
\label{assump:var_inter}
There exists a hyperbox  $$\mathcal{C}_z\coloneqq[-C_{z,1}, C_{z, 1}]\times\cdots\times[-C_{z,m}, C_{z, m}]$$ with $C_{z,i}>0,i\in[m]$, such that $\forall x\in\mathcal{X}$, the corresponding intermediate variables $z\in\mathcal{C}_z$. Furthermore, we choose $C_{z,i+1}\geq B_i,i\in\IB$. 
\end{assumption}

To facilitate the following discussion, we use shorthand $$\mathcal{S}_i=\mathcal{X}\times[-C_{z,1}, C_{z,1}]\times\cdots\times[-C_{z,i}, C_{z,i}].$$ 
\begin{assumption}
\label{assump:Lip}
The functions $\phi_i$, $i\in\{0,\cdots,m-1\}$ are Lipschitz continuous with respect to one norm over $\mathcal{S}_i\coloneqq\mathcal{X}\times[-C_{z,1}, C_{z,1}]\times\cdots\times[-C_{z,i}, C_{z,i}]$. That is to say, there exist constants $L_{\phi_i}>0$, such that $\forall s_i^1, s_i^2\in\mathcal{S}_i$, we have, 
\begin{equation}
  \|\phi_i(s_i^1)-\phi_i(s_i^2)\|\leq L_{\phi_i}\|s_i^1-s_i^2\|_1,  
\end{equation} 
\end{assumption}
\begin{assumption}
\label{assump:eval_model}
The black-box functions $\phi_i,\forall i\in\IB$ can be separately evaluated. That is to say we get access to an oracle to evaluate each black-box function $\phi_i$, $i\in\IB$. Furthermore, each observation of the output for the black-box functions is corrupted by independent identically distributed $\sigma$-sub-Gaussian noise.~\footnote{Note that the noise is only for observation. There is no process noise in the compositional chain of the grey-box function.}   
\end{assumption}




\section{Learning of the black-box functions with Gaussian process} 
We use the independent Gaussian processes to learn the unknown black-box functions $\phi_i,i\in\IB$. Similar to~\cite{chowdhury2017kernelized}, we artificially introduce a Gaussian process $\mathcal{GP}(0, k_i(\cdot, \cdot))$ for the surrogate modeling of the unknown black-box function $\phi_i$. We also adopt an \emph{i.i.d} Gaussian zero-mean noise model with noise variance $\lambda>0$.
\begin{myRemark}
Note that this Gaussian process model is \emph{only} used to derive posterior mean functions, covariance functions, and maximum information gain for the purpose of algorithm design. We introduce the artificial Gaussian process only to make the algorithm more interpretable. It does not change our set-up that $\phi_i,i\in\IB$ is a deterministic function and that the observational noise only needs to be \emph{i.i.d} $\sigma$-sub-Gaussian. Indeed, we are considering the agnostic setting introduced in~\cite{srinivas2012information} and the noise $\lambda$ here can also be interpreted as a regularizer that stabilizes the numerical computation.
\end{myRemark}
We use $(\bar{s}_i^t, \tilde{z}^t_{i+1}),i\in\IB$ to denote the data we add to learn the black-box function $\phi_i$ at step $t$, where $\tilde{z}^t_{i+1}=\phi_i(\bar{s}_i^t)+\bar{\xi}_i^t$. We introduce the following functions for $s_i, s_i^{\prime}$ for the learning of the function $\phi_i, i\in\IB$,
\bse
\label{eq:mean_cov}
\begin{align}
\mu_{i,t}(s_i) &=k_{i}(\bar{s}_i^{1:t}, s_i)^\top\left(K_{i,t}+\lambda I\right)^{-1} \tilde{z}_{i+1}^{1:t}\enspace, \\
k_{i,t}\left(s_i, s_i^{\prime}\right) &=k_{i}\left(s_i, s_i^{\prime}\right)-k_{i}(\bar{s}_i^{1:t}, s_i)^\top\left(K_{i,t}+ \lambda I\right)^{-1} k_{i}\left(\bar{s}_i^{1:t}, s_i^{\prime}\right), \\
\sigma_{i,t}^2(s_i) &=k_{i,t}(s_i, s_i)\enspace
\end{align}
\ese
with $k_{i}(\bar{s}_i^{1:t}, s_i)=[k_i(\bar{s}_i^1, s_i), k_i(\bar{s}_i^2, s_i),\cdots, k_i(\bar{s}_i^t, s_i)]^\top$, $K_{i,t}=(k_i(\bar{s}^{\tau_1}_i,\bar{s}^{\tau_2}_i))_{{\tau_1},{\tau_2}\in[t]}$ and $\tilde{z}_{i+1}^{1:t}=[\tilde{z}_{i+1}^1, \tilde{z}_{i+1}^2,\cdots,\tilde{z}_{i+1}^t]^\top$. We also introduce the maximum information gain for the function $\phi_i$ as proposed in~\cite{srinivas2012information},
\bee
\label{eq:max_inf_gain}
\gamma_{i,t}:=\max_{S_i \subset \mathcal{S}_i ;|S_i|=t} \frac{1}{2} \log \left|I+\lambda^{-1}K_{i,S_i}\right|,
\eee
where $K_{i,S_i}=(k_i(s_i,s_i^\prime))_{s_i,s_i^\prime\in S_i}$. 
\begin{lemma}[Theorem 2,~\cite{chowdhury2017kernelized}]
Let Assumptions~\ref{assump:support_set}, \ref{assump:bounded_norm}, \ref{assump:regular_kernel}, \ref{assump:var_inter} and \ref{assump:eval_model} hold. For any $\delta\in(0, 1)$, with probability at least $1-\delta/m$, the following holds for all $s_i \in \mathcal{S}_i,i\in\IB$ and $1\leq t \leq T$, $T\in\mathbb{N}$, 
\begin{align}
\left|\mu_{i,t-1}(s_i)-\phi_i(s_i)\right|\leq\left(B_i+\sigma \sqrt{2\left(\gamma_{i,t-1}+1+\ln (m / \delta)\right)}\right) \sigma_{i,t-1}(s_i),
\end{align}
where $\mu_{i,t-1}(s_i), \sigma^2_{i,t-1}(x)$ and $\gamma_{i,t-1}$ are as given in Eq.~\eqref{eq:mean_cov} and Eq.~\eqref{eq:max_inf_gain}, and $\lambda$ is set to be $1+\frac{2}{T}$.~($\mu_{i,0}$ and $\sigma_{i,0}$ are the prior mean and standard deviation, and $\gamma_{i,0}=0$.) \label{lem:conf_int} 
\end{lemma}
\begin{myRemark}
Note that we replace the `$\delta$' in Theorem 2 of~\cite{chowdhury2017kernelized} by $\delta/m$, which will be useful to derive the confidence interval with $1-\delta$ probability in Lem.~\ref{lem:hpb}.     
\end{myRemark}

To facilitate the following algorithm design and discussion, we introduce the lower confidence and upper confidence bound functions,
\begin{definition}
For $i\in\IB$, $s_i\in\mathcal{S}_i$ and $t\in[T]$, 
\bse
\begin{align}
l_{i,t}(s_i)&\defeq\max\left\{\mu_{i,t-1}(s_i)-\betac_{i,t}\sigma_{i,t-1}(s_i), -B_i\right\}\enspace,\\
u_{i,t}(s_i)&\defeq\min\left\{\mu_{i,t-1}(s_i)+\betac_{i,t}\sigma_{i,t-1}(s_i), B_i\right\}
\end{align}
\ese
with $\betac_{i,t}=B_i+\sigma \sqrt{2\left(\gamma_{i,t-1}+1+\ln (m / \delta)\right)}$.
\label{def:hpb_int}
\end{definition}
Lem.~\ref{lem:hpb} then follows, 
\begin{lemma}[Corollary 2.6,~\cite{xu2022constrained}]
    \label{lem:hpb}
Let Assumptions~\ref{assump:support_set}, \ref{assump:bounded_norm}, \ref{assump:regular_kernel}, \ref{assump:var_inter} and \ref{assump:eval_model} hold.
With probability at least $1-\delta$, the following holds for all $s_i\in\mathcal{S}_i, i\in\IB$ and $1\leq t\leq T$,
\bee
\forall i\in\IB,\;\;
\phi_i(s_i)&\in[l_{i,t}(s_i),u_{i,t}(s_i)]\enspace. 
\eee
\end{lemma}

\section{Algorithm}
 The Bayesian optimization algorithm of expensive nested grey-box functions is given as in Alg.~\ref{alg:grey_box_opt}. 
\begin{algorithm}[htbp!]
\caption{Bayesian Optimization of Expensive Nested {Grey} Box Functions}
\label{alg:grey_box_opt}
\begin{algorithmic}[1]
\FOR{$t\in[T]$}
\STATE Find $(x^t, \bar{z}^t)$ by solving
\vspace{-0.2cm}
\begin{subequations}
\begin{align}
    \min_{x\in \mathcal{X}, z\in\mathbb{R}^m}&\;\; z_m\\
    \text{ subject to }&\;\; l_{i,t}(s_i)\leq z_{i+1}\leq u_{i,t}(s_i),\forall i\in\IB,\\
    &\;\;z_{i+1}=\phi_i(s_i), \forall i\in\IW,\\
    &\;\;s_i=[x_1, \cdots, x_n,z_1,\cdots, z_i]^\top, \forall i\in\{0\}\cup[m].
\end{align}
\end{subequations}
\vspace{-0.3cm}
\label{alg_line:aux_prob}
\STATE Evaluate the grey-box function $f$ at $x^t$, incur intermediate variables $z^t$ and final objective  $f(x^t)=z^t_m$.
\STATE Separately get the noisy evaluation $\tilde{z}_{i+1}^t\coloneqq\phi_i(\bar{s}_i^t)+\bar{\xi}_{i,t}$ of the black-box function $\phi_i,i\in\IB$ at the point $\bar{s}^t_i\coloneqq[x^t_1, \cdots, x^t_n,\bar{z}^t_1,\cdots,\bar{z}^t_i]^\top$.~\label{alg_line:sample_point}  
\STATE Update Gaussian process posterior mean and covariance for each black-box function with the new evaluations added. 
\ENDFOR
\end{algorithmic}
\end{algorithm}
The idea of the Alg.~\ref{alg:grey_box_opt} is that in each step, we solve an optimistic auxiliary problem of the original problem where we require the augmented states $z_{i+1}$, $i\in\IB$ need to be inside the confidence interval conditioned on the current observation history.~\footnote{Note that to observe these intermediate variables, we may suffer from observational noise, which may introduce noise perturbation to the subsequent Gaussian process input.}   
\begin{myRemark}
   We assume that an oracle can efficiently solve the auxiliary problem in line~\ref{alg_line:aux_prob}. In low dimensions, we can use pure grid search to solve it. In higher dimensional cases, we can apply a gradient-based optimization algorithm with different starting points. 
\end{myRemark}

\section{Analysis}
We now analyze the performance of the algorithm. We use $x^*$ to denote the optimal solution of the Prob.~\eqref{eqn:prob_form} and $z^*$ to denote the corresponding intermediate variables. We are interested in bounding the cumulative regret, 
\begin{equation}
    R_T=\sum_{t=1}^{T}\left(f(x^t)-f(x^*)\right).
\end{equation}
We first introduce a lemma to bound the discrepancy between the real output of a grey box function and the \emph{possible} output of the grey-box function that is consistent with the posterior of Gaussian process with high probability.
\begin{lemma}[Output Discrepancy Bound]
\label{lem:discrepancy_bound} 
Let assumptions \ref{assump:support_set}, \ref{assump:bounded_norm}, \ref{assump:regular_kernel}, \ref{assump:var_inter}, \ref{assump:Lip}, and \ref{assump:eval_model} hold. For any $x^t\in\mathcal{X}$, let $z^t$ denote its corresponding augmented state when evaluating the grey-box function at $x^t$. Then for any $z$ that satisfies, 
\begin{align}
    l_{i,t}(s_i)\leq z_{i+1}\leq u_{i,t}(s_i),\;\;&\forall i\in\IB,\\
    z_{i+1}=\phi_i(s_i),\;\;& \forall i\in\IW
\end{align}
with $s_i=[x^t_1, \cdots, x^t_n,z_1,\cdots, z_i]^\top, \forall i\in\{0\}\cup[m-1]$. We have, with probability at least $1-\delta$,  
$$|f(x^t)-z_m|=|z^t_m-z_m|\leq\sum_{i\in\IB}A_i\beta_{i,t}\sigma_{i,t-1}({s}_{i})$$
with $A_i=2\left(\sum_{j=1}^{m-1-i}2^j\prod_{i<k_1<\cdots<k_j= m-1}L_{\phi_{k_j}} \right)$.
\end{lemma}
\begin{proof}
We consider bounding $|z_{i+1}^t-{z}_{i+1}|,i\in[m-1]$ recursively.\\
\textbf{Case 1:} $\phi_i$ is a known white-box function, then,
\begin{equation}
    |z_{i+1}^t-{z}_{i+1}|=|\phi_i(s_i^t)-\phi_i({s}_i)|\leq L_{\phi_i}\|s_i^t-{s}_i\|_1\leq 2L_{\phi_i}\sum_{j=1}^i|z^t_j-{z}_j|.
\end{equation}
\textbf{Case 2:} $\phi_i$ is an unknown black-box function, then,
\bse
\begin{align}
&|z_{i+1}^t-{z}_{i+1}|\\
\leq\;&|\phi_i(s_i^t)-l_{i,t}({s}_i)|+|\phi_i(s_i^t)-u_{i,t}({s}_i)|\label{inq:int}\\\notag
\leq\;&|\phi_i(s_i^t)-\phi_i({s}_i)|+|\phi_i({s}_i)-l_{i,t}({s}_i)|\\
&+|\phi_i(s_i^t)-\phi_i({s}_i)|+|\phi_i({s}_i)-u_{i,t}({s}_i)|\label{inq:tri}\\
\leq\;& 2L_{\phi_i}\|s_i^t-{s}_i\|+2\beta_{i,t}\sigma_{i,t-1}({s}_i)\label{inq:lip}\\
=\;&2L_{\phi_i}\sum_{j=1}^i|z_j^t-{z}_j|+2\beta_{i,t}\sigma_{i,t-1}({s}_i).
\end{align}
\ese
where the inequality~\eqref{inq:int} follows by $z^t_{i+1}=\phi_i(z^t_i)$ and ${z}_{i+1}\in[l_{i,t}({s}_i),u_{i,t}({s}_i)]$, the inequality~\eqref{inq:tri} follows by the triangle inequality, and the inequality~\eqref{inq:lip} follows by the Lipschitz assumption and that ${z}_{i+1}\in[l_{i,t}({s}_i),u_{i,t}({s}_i)]$.   
Solving the recursive inequalities above, we get
\bse
\begin{align}
&|z_m^t-{z}_m|\\
\leq&\sum_{i\in\IB}2\left(\sum_{j=1}^{m-1-i}2^j\prod_{i<k_1<\cdots<k_j= m-1}L_{\phi_{k_j}} \right)\beta_{i,t}\sigma_{i,t-1}({s}_i),
\end{align}
\ese
which concludes the proof. \hfill $\square$
\end{proof}
We then have a lemma to bound the single-step instantaneous regret. 
\begin{lemma}
\label{lem:inst_r}
 Let assumptions \ref{assump:support_set}, \ref{assump:bounded_norm}, \ref{assump:regular_kernel}, \ref{assump:var_inter}, \ref{assump:Lip}, and \ref{assump:eval_model} hold. There exists constants $A_{i}, i\in\IB$, such that with probability at least $1-\delta$, 
    \begin{equation}
        [f(x^t)-f(x^*)]^+\leq\sum_{i\in\IB}A_i\beta_{i,t}\sigma_{i,t-1}(\bar{s}_{i}^t)
    \end{equation}
    with $A_i=2\left(\sum_{j=1}^{m-1-i}2^j\prod_{i<k_1<\cdots<k_j= m-1}L_{\phi_{k_j}} \right)$. 
\end{lemma}
\begin{proof}
With probability at least $1-\delta$,
\bse
\begin{align}
&[f(x^t)-f(x^*)]^+\\
=\;&[z_m^t-\bar{z}_m^t+\bar{z}_m^t-f(x^*)]^+\label{pf_line:zeq}\\
\leq\;&[z_m^t-\bar{z}_m^t]^++[\bar{z}_m^t-f(x^*)]^+\label{pf_line:split_pos}\\
\leq\;&[z_m^t-\bar{z}_m^t]^+\label{pf_line:opt_barz}
\end{align}
\ese
where the equality~\eqref{pf_line:zeq} follows by the definition of $z^t$, the inequality~\eqref{pf_line:opt_barz} follows by the optimality of $(x^t,\bar{z}^t)$ for the auxiliary problem in line~\ref{alg_line:aux_prob} of the Alg.~\ref{alg:grey_box_opt} and the feasibility of $(x^*, z^*)$ for the same auxiliary problem. 
Applying the bound in Lemma~\ref{lem:discrepancy_bound} and setting $A_i=2\left(\sum_{j=1}^{m-1-i}2^j\prod_{i<k_1<\cdots<k_j= m-1}L_{\phi_{k_j}} \right)$ concludes the proof. \hfill$\square$
\end{proof}

We then restate a useful lemma to bound the cumulative posterior standard deviation along the sampling trajectory, 
\begin{lemma}[Lemma 4,~\cite{chowdhury2017kernelized_arxiv}]
If $\bar{s}^1_i, \bar{s}^2_i, \cdots, \bar{s}^T_i$ are the points sampled by Alg.~\ref{alg:grey_box_opt} in line~\ref{alg_line:sample_point}, then,
\bee
\label{lem:bound_cumu_sd}
\sum_{t=1}^T \sigma_{i, t-1}\left(\bar{s}^t_i\right) \leq \sqrt{4(T+2) \gamma_{i,T}}.
\eee
\end{lemma}
\begin{theorem}
\label{thm:R_bound}
Let assumptions \ref{assump:support_set}, \ref{assump:bounded_norm}, \ref{assump:regular_kernel}, \ref{assump:var_inter}, \ref{assump:Lip}, and \ref{assump:eval_model} hold. Then, with probability at least $1-\delta$, the sample points of Algorithm \ref{alg:grey_box_opt} satisfy, 
\bee
R_T&\leq \sum_{i\in\IB}2A_i\betac_{i,T}\sqrt{(T+2)\gamma_{i,T}}=\mathcal{O}(\sum_{i\in\IB}A_i\gamma_{i,T}\sqrt{T})\enspace,
\eee
where $A_i$ is a Lipschitz constant dependent constant as given in Lem.~\ref{lem:inst_r}.
\end{theorem}
\begin{proof}
We have 
\bse
\begin{align}
    R_T&=\sum_{t=1}^T(f(x^t)-f(x^*))\\
    &\leq\sum_{t=1}^T\sum_{i\in\IB}A_i\beta_{i,t}\sigma_{i,t-1}(\bar{s}_{i}^t)~\label{inq:single_step}\\
    &\leq\sum_{t=1}^T\sum_{i\in\IB}A_i\beta_{i,T}\sigma_{i,t-1}(\bar{s}_{i}^t)~\label{inq:mono}\\
    &=\sum_{i\in\IB}A_i\beta_{i,T}\sum_{t=1}^T\sigma_{i,t-1}(\bar{s}_{i}^t)\\ &\leq\sum_{i\in\IB}A_i\beta_{i,T}\sqrt{4(T+2)\gamma_{i,T}},~\label{inq:bound_cumu_sigma}
\end{align}
\ese
where the inequality \eqref{inq:single_step} follows by Lem.~\ref{lem:inst_r}, the inequality \eqref{inq:mono} follows by the monotonicity of $\beta_{i,t}$, and the inequality \eqref{inq:bound_cumu_sigma} follows by Lem.~\ref{lem:bound_cumu_sd}. Plugging in the definition of $\beta_{i,T}$ completes the proof.    
\hfill$\square$
\end{proof}

The best iterate convergence guarantee can be obtained, consequently given below.
\begin{theorem}
\label{thm:best_ite_conv}
Let assumptions \ref{assump:support_set}, \ref{assump:bounded_norm}, \ref{assump:regular_kernel}, \ref{assump:var_inter}, \ref{assump:Lip}, and \ref{assump:eval_model} hold. With probability at least $1-\delta$, there exists $\tilde{x}^T\in\{x^1,\cdots, x^T\}$, such that,  
\[
f(\tilde{x}^T)-f(x^*)\leq\mathcal{O}\left(\frac{\sum_{i\in\IB}A_i\gamma_{i,T}}{\sqrt{T}}\right) \enspace.
\]
\end{theorem}
\begin{proof}
   By Thm.~\ref{thm:R_bound}, we get
 \bse
\begin{align}
    R_T&=\sum_{t=1}^T(f(x^t)-f(x^*))\\
    &\leq\mathcal{O}\left(\sum_{i\in\IB}A_i\gamma_{i,T}\sqrt{T}\right).
\end{align}
\ese
Therefore,
\begin{subequations}
\begin{align}    
    \frac{R_T}{T}&=\sum_{t=1}^T\frac{f(x^t)-f(x^*)}{T}\\
    &\leq\mathcal{O}\left(\frac{\sum_{i\in\IB}A_i\gamma_{i,T}}{\sqrt{T}}\right).
\end{align}
\end{subequations}
Since $f(x^t)-f(x^*)\geq0$, there exists $\tilde{x}^T\in\{x^1, \cdots,x^T\}$, such that,
\begin{align}    
f(\tilde{x}^T)-f(x^*)\leq\frac{R_T}{T}\leq\mathcal{O}\left(\frac{\sum_{i\in\IB}A_i\gamma_{i,T}}{\sqrt{T}}\right)
\end{align}
concludes the proof. \hfill$\square$
\end{proof}

If all the kernel functions corresponding to different black-box functions are of the same type, we can list the convergence rate as in the Tab.~\ref{tab:kern_spec_bounds} by applying the bounds on maximum information gains from~\cite{srinivas2012information,vakili2021information}.
\begin{table*}[htbp!]
\caption{Kernel-specific cumulative regret bounds and the convergence rate to the optimal solution. In the table, $d=n+m$ and $\nu$ represent the smoothness parameter of the M\'atern kernel.} 
\label{tab:kern_spec_bounds}
\renewcommand{\arraystretch}{1.5}
\centering
\resizebox{0.99\columnwidth}{!}{
\begin{tabular}{|c|c|c|c|c|}
\hline
\textbf{Kernel} & \textbf{Maximum Information Gain}& \textbf{Cumulative Regret} & \textbf{Convergence Rate} \\
\hline 
\textbf{Linear} &$\mathcal{O}(d\log T)$ &$\mathcal{O}\left(d\log T\sqrt{T}\right)$  &$\mathcal{O}\left(\frac{d\log T}{\sqrt{T}}\right)$  \\
\hline
\begin{tabular}{@{}c@{}}
\textbf{Squared }\\[-0.1cm]
\textbf{Exponential}
\end{tabular}
& $\mathcal{O}((\log T)^{d+1})$ &$\mathcal{O}\left((\log T)^{d+1}\sqrt{T}\right)$ &$\mathcal{O}\left(\frac{(\log T)^{d+1}}{\sqrt{T}}\right)$  \\
\hline 
\begin{tabular}{@{}c@{}}
\textbf{M\'atern}\\[-0.1cm]
{$\left(\nu>\frac{d}{2}\right)$}
\end{tabular}
&{$\mathcal{O}(T^{\frac{d }{2 \nu+d}}\log^{\frac{2\nu}{2\nu+d}}(T))$} & {$\mathcal{O}(T^{\frac{2\nu+3d }{4 \nu+2d}}\log^{\frac{2\nu}{2\nu+d}}(T))$}  &{$\mathcal{O}\left(T^{-\frac{2\nu-d}{4 \nu+2d}}\log^{\frac{2\nu}{2\nu+d}}(T)\right)$}   \\ 
\hline
\end{tabular}
}
\end{table*}
\vspace{-0.5cm}
\section{Extension to the Constrained Case}
\label{sec:ext_cons}
We have given the results in the unconstrained case. Following the same idea as in~\cite{xu2022constrained}, we can extend the Alg.~\ref{alg:grey_box_opt} to the constrained case and derive bounds on the cumulative violations. We formulate our problem as in~\ref{eqn:prob_form},
\begin{subequations}
\label{eqn:constrained_prob_form}
\begin{eqnarray}
\min_{x\in\mathcal{X}}\quad &&f(x),\label{eqn:constrained_form}\\
\textrm{subject to}&& g_\citer(x)\leq 0, \citer\in[\tcon],
\end{eqnarray}
\end{subequations}
where $g_\citer, \citer\in[\tcon]$ are also grey-box functions modeled in the same way as on the objective function $f$. We make the same regularity assumptions~\ref{assump:bounded_norm},~\ref{assump:regular_kernel},~\ref{assump:var_inter},~\ref{assump:Lip} and \ref{assump:feas} on $g_k,k\in[K]$ as on the objective function $f$. To distinguish the grey-box modeling notations for different functions, we add a superscript of function. For example, we use $z_i^{t, g_\citer}, i\in[m^{g_\citer}]$ to denote the $i$-th intermediate variable for the grey-box function $g_\citer$ at step $t$. 
In this section, we first focus on the feasible case and introduce the feasibility assumption. We will later show an infeasibility detection scheme in Sec.~\ref{sec:inf_dec}.  
\begin{assumption}
    \label{assump:feas} 
    There exists $\tilde{x}\in\mathcal{X}$, such that $g_\citer(\tilde{x})\leq0,\forall\citer\in[\tcon]$.
\end{assumption}
To make the algorithmic description more compact, the following plausible set definition is introduced.

\begin{definition}[Plausible set of $z$]
Conditioned on the historical samples up to step $t$, the plausible set $\mathcal{Z}_t^f(x)$ of $z^f$ with input $x$ is defined as the feasible set of the following feasibility problem,
\begin{subequations}
\begin{align}
\min_{z^f\in\mathbb{R}^{m^f}}\;\; &1\\
\text{subject to}\quad&l_{i,t}(s^f_i)\leq z^f_{i+1}\leq u_{i,t}(s^f_i),\forall i\in\IB^f,\\
&z^f_{i+1}=\phi^f_i(s^f_i), \forall i\in\IW^f,\label{alg_line:cons_white_box}\\
&s^f_i=[x_1, \cdots, x_n,z^f_1,\cdots, z^f_i]^\top, \forall i\in\{0\}\cup[m^f-1].
\end{align}
\end{subequations}
\end{definition}
The plausible set corresponding to $g_k$ can be defined in the same way.
\begin{algorithm}[htbp!]
\caption{Constrained Bayesian Optimization of Expensive Nested {Grey} Box Functions}
\label{alg:constrained_grey_box_opt}
\begin{algorithmic}[1]
\normalsize
\FOR{$t\in[T]$}
\STATE Find $(x^t, \bar{z}^{t,f}, (\bar{z}^{t,g_\citer})_{\citer\in[\tcon]})$ by solving
\vspace{-0.2cm}
\begin{subequations}
\begin{align}
\min_{x\in \mathcal{X}, z^f\in\mathbb{R}^{m^f},z^{g_\citer}\in\mathbb{R}^{m^{g_\citer}}, \citer\in[\tcon]}\;\; &z^f_{m^f} \\
\text{subject to}\qquad\qquad &z^f\in\mathcal{Z}_t^f(x),\\
\;\;&z^{g_k}\in\mathcal{Z}_t^{g_k}(x), \forall\citer\in[\tcon]\\
\;\;&z^{g_\citer}_{m^{g_\citer}}\leq0, \forall\citer\in[\tcon]    
\end{align}
\end{subequations}
\vspace{-0.3cm}
\label{alg_line:aux_prob}
\label{alg_line:constrained_aux_prob}
\STATE Evaluate the grey-box function $f$ and $g_\citer,\citer\in[\tcon]$ at $x^t$, incur objective $f(x^t)$, and constraints $g_\citer(x^t),\citer\in[\tcon]$ with the input $x^t$.
\STATE Separately evaluate the black-box function $\phi^f_i,i\in\IB^f$ at the point $\bar{s}^{t,f}_i\coloneqq[x^{t}_1, \cdots, x^{t}_n,\bar{z}^{t,f}_1,\cdots,\bar{z}^{t,f}_i]^\top$. Also separately evaluate the black-box function $\phi^{g_\citer}_i,i\in\IB^{g_\citer}$ at the point $\bar{s}^{t,g_\citer}_i\coloneqq[x^{t}_1, \cdots, x^{t}_n,\bar{z}^{t,g_\citer}_1,\cdots,\bar{z}^{t,g_\citer}_i]^\top$.  
\STATE Update Gaussian process posterior mean and covariance for each black-box function with the new evaluations added.
\ENDFOR
\end{algorithmic}
\end{algorithm}
The observation~\ref{obs:feas} guarantees the feasibility of the auxiliary problem in Alg.~\ref{alg:constrained_grey_box_opt}.
\begin{obs}
Since $x^\star$ is a feasible point for the original constrained optimization problem, then with high probability, $x^\star$ with the corresponding $z^{\star, f}$ and $z^{\star, g_k}, k\in[K]$ is a feasible solution for the auxiliary optimization problem in Alg.~\ref{alg:constrained_grey_box_opt}. 
\label{obs:feas}
\end{obs}

For the constrained problem, violations need to be taken care of in addition to the cumulative regret~\cite{xu2022constrained}. We formally define cumulative violations up to step $T$ as,
\begin{definition}
   Cumulative violation up to step $T$ is defined as,
  \begin{equation}
     \mathcal{V}_{\citer, T}\coloneqq \sum_{t=1}^T[g_\citer(x_t)]^+. 
  \end{equation} 
\end{definition}
We then have a theorem to bound both the cumulative regret and the cumulative violations.
\begin{theorem}
\label{thm:RV_bound}
Let assumptions \ref{assump:support_set}, \ref{assump:bounded_norm}, \ref{assump:regular_kernel}, \ref{assump:var_inter}, \ref{assump:Lip}, \ref{assump:eval_model} and \ref{assump:feas} hold. We have, with probability at least $1-\delta$, the sample points of Algorithm \ref{alg:constrained_grey_box_opt} satisfy, 
\begin{subequations}
\begin{align}\notag
R_T&\leq\sum_{t=1}^T[f(x^t)-f(x^*)]^+\\
&\leq \sum_{i\in\IB^f}2A_i^f\betac^f_{i,T}\sqrt{(T+2)\gamma^f_{i,T}}=\mathcal{O}\left(\sum_{i\in\IB^f}A^f_i\gamma^f_{i,T}\sqrt{T}\right)\enspace,\\
\mathcal{V}_{\citer,T}&\leq \sum_{i\in\IB^{g_\citer}}2A_i^{g_\citer}\betac^{g_\citer}_{i,T}\sqrt{(T+2)\gamma^{g_\citer}_{i,T}}\\
&=\mathcal{O}\left(\sum_{i\in\IB^{g_\citer}}A^{g_\citer}_i\gamma^{g_\citer}_{i,T}\sqrt{T}\right)\enspace,\forall\citer\in[\tcon]
\end{align}
\end{subequations}
where $A^f_i$ and $A^{g_\citer}_i$ are Lipschitz constant dependent constants.
\end{theorem}
\begin{proof}
We have
\begin{subequations}
\begin{align}
 R_T=&\sum_{t=1}^T(f(x^t)-f(x^*))\\
    \leq&\sum_{t=1}^T[f(x^t)-f(x^*)]^+\\
    \leq&\sum_{t=1}^T\sum_{i\in\IB^f}A^f_i\beta^f_{i,t}\sigma_{i,t-1}(\bar{s}_{i}^{t,f})~\label{inq:cons_inst_r}\\
    \leq&\sum_{i\in\IB^f}A^f_i\beta^f_{i,T}\sum_{t=1}^T\sigma_{i,t-1}(\bar{s}_{i}^{t,f})\\~\label{inq:cons_bound_cumu_sigma}     \leq&\sum_{i\in\IB^f}A^f_i\beta^f_{i,T}\sqrt{4(T+2)\gamma^f_{i,T}},
\end{align}
\end{subequations}
where the inequality~\eqref{inq:cons_inst_r} follows by the Lem.~\ref{lem:discrepancy_bound} since $(x^*, z^{*,f}, (z^{*,g_k})_{k\in[K]})$ is a feasible solution for the constrained auxiliary optimization problem in line~\ref{alg_line:constrained_aux_prob} of the Alg.~\ref{alg:constrained_grey_box_opt}. 

Similarly, we have
\begin{subequations}
\begin{align}
 \mathcal{V}_{k,T}=&\sum_{t=1}^T[g_k(x_t)]^+\\
    \leq&\sum_{t=1}^T[g_k(x^t)-\bar{z}_{m^{g_k}}^{t,g_k}]^++[\bar{z}_{m^{g_k}}^{t,g_k}]^+~\label{inq:inst_r}\\
    =&\sum_{t=1}^T [g_k(x^t)-\bar{z}_{m^{g_k}}^{t,g_k}]^+~\label{eq:remove_neg}\\   \leq&\sum_{i\in\IB^{g_k}}A^{g_k}_i\beta^{g_k}_{i,T}\sqrt{4(T+2)\gamma^{g_k}_{i,T}},~\label{inq:cons_discrep}
\end{align}
\end{subequations}
where the equality \eqref{eq:remove_neg} follows by the feasibility of $\bar{z}^{t,g_k}$ for the auxiliary constrained optimization problem and the inequality \eqref{inq:cons_discrep} follows by the discrepancy bound in Lem.~\ref{lem:discrepancy_bound}.  \hfill$\square$
\end{proof}

Consequently, we can derive a convergence rate to the constrained global optimum.    
\begin{theorem}
\label{thm:converge_constrained_opt}
Let assumptions \ref{assump:support_set}, \ref{assump:bounded_norm}, \ref{assump:regular_kernel}, \ref{assump:var_inter}, \ref{assump:Lip}, \ref{assump:eval_model} and \ref{assump:feas} hold. We have, with probability at least $1-\delta$, there exists $\tilde{x}_T$ from the sample points of Algorithm \ref{alg:constrained_grey_box_opt} satisfying, 
\bse
\begin{align}
f(\tilde{x}_T)-f(x^*)&\leq\mathcal{O}\left(\frac{\sum_{i\in\IB^f}A^f_i\gamma^f_{i,T}+\sum_{\citer\in[\tcon]}\sum_{i\in\IB^{g_\citer}}A^{g_\citer}_i\gamma^{g_\citer}_{i,T}}{\sqrt{T}}\right)\enspace,\\
[g_\citer(\tilde{x}_T)]^+&\leq \mathcal{O}\left(\frac{\sum_{i\in\IB^f}A^f_i\gamma^f_{i,T}+\sum_{\citer\in[\tcon]}\sum_{i\in\IB^{g_\citer}}A^{g_\citer}_i\gamma^{g_\citer}_{i,T}}{\sqrt{T}}\right)\enspace,~\forall\citer\in[\tcon]
\end{align}
\ese
where $A^f_i$ and $A^{g_\citer}_i$ are constants dependent on the Lipschitz constants.
\end{theorem}
\begin{proof}
From theorem~\ref{thm:RV_bound}, we can derive, 
\begin{subequations}
\begin{align}
\sum_{t=1}^T[f(x^t)-f(x^*)]^+\leq& \sum_{i\in\IB^f}2A_i^f\betac^f_{i,T}\sqrt{(T+2)\gamma^f_{i,T}}\enspace,\\
\sum_{t=1}^T[g_\citer(x^t)]^+\leq& \sum_{i\in\IB^{g_\citer}}2A_i^{g_\citer}\betac^{g_\citer}_{i,T}\sqrt{(T+2)\gamma^{g_\citer}_{i,T}},~\forall\citer\in[\tcon]\enspace.
\end{align}
\end{subequations}
Summing up the above inequalities and dividing by $T$, we get,
\begin{subequations}
\begin{align}
&\frac{1}{T}\sum_{t=1}^T\left([f(x^t)-f(x^*)]^++\sum_{k=1}^\tcon[g_\citer(x^t)]^+\right)\\
\leq\;&\frac{\sum_{i\in\IB^f}2A_i^f\betac^f_{i,T}\sqrt{(T+2)\gamma^f_{i,T}}+\sum_{\citer=1}^\tcon\sum_{i\in\IB^{g_\citer}}2A_i^{g_\citer}\betac^{g_\citer}_{i,T}\sqrt{(T+2)\gamma^{g_\citer}_{i,T}}}{T}\\
=\;&\mathcal{O}\left(\frac{\sum_{i\in\IB^f}A^f_i\gamma^f_{i,T}+\sum_{\citer\in[\tcon]}\sum_{i\in\IB^{g_\citer}}A^{g_\citer}_i\gamma^{g_\citer}_{i,T}}{\sqrt{T}}\right)\enspace.
\end{align}
\end{subequations}
Since $[f(x^t)-f(x^*)]^++\sum_{\citer=1}^\tcon[g_\citer(x^t)]^+\geq0$, there exists $\tilde{x}^T\in\{x^1, \cdots,x^T\}$, such that,
\begin{subequations}
\begin{align}
&[f(\tilde{x}^T)-f(x^*)]^++\sum_{k=1}^\tcon[g_\citer(\tilde{x}^T)]^+\\
\leq\;&\frac{1}{T}\sum_{t=1}^T\left([f(x^t)-f(x^*)]^++\sum_{k=1}^\tcon[g_\citer(x^t)]^+\right)\\
=\;&\mathcal{O}\left(\frac{\sum_{i\in\IB^f}A^f_i\gamma^f_{i,T}+\sum_{\citer\in[\tcon]}\sum_{i\in\IB^{g_\citer}}A^{g_\citer}_i\gamma^{g_\citer}_{i,T}}{\sqrt{T}}\right)\enspace.
\end{align}
\end{subequations}
The desired results then follow. \hfill $\square$
\end{proof}

\subsection{Infeasibility Detection}
\label{sec:inf_dec}
The cumulative regret/violation bounds and convergence results are derived under the assumption that the problem is feasible. However, in practice, the grey-box problem may indeed be infeasible and therefore, an infeasibility detection scheme is highly desired. In the following, we show that if the original problem is infeasible, we will find that the auxiliary problem in line~\ref{alg_line:aux_prob} of Alg.~\ref{alg:constrained_grey_box_opt} infeasible within finite steps. That is to say, we are able to detect the infeasibility of the original problem through the detection of the infeasibility of the optimistic auxiliary problem.                 

For infeasibility detection, we try to avoid `false positives', where a feasible problem is declared as infeasible, and `false negatives', where an infeasible problem is not declared to be infeasible. The observation~\ref{obs:feas} shows that it is impossible to have `false positives'. The following theorem complements the observation~\ref{obs:feas} by showing that if the original problem is infeasible, our algorithm is guaranteed to declare infeasibility within finite steps with high probability.  
\begin{theorem}
Let assumptions \ref{assump:support_set}, \ref{assump:bounded_norm}, \ref{assump:regular_kernel}, \ref{assump:var_inter}, \ref{assump:Lip}, and \ref{assump:eval_model} hold. Assume that Prob.~\eqref{eqn:prob_form} is infeasible, that is, there exists $\citer\in[\tcon]$, such that 
\[
\min_{x\in\mathcal{X}}\;\;g_k(x)=\epsilon>0,
\] 
and $\lim_{T\to\infty}\frac{\sum_{i\in\IB^{g_\citer}}A^{g_\citer}_i\gamma^{g_\citer}_{i,T}}{\sqrt{T}}=0$. Given a desired confidence level $\delta\in(0,1)$, the auxiliary problem in line~\ref{alg_line:aux_prob} of the Alg.~\ref{alg:constrained_grey_box_opt} 
will be found infeasible within a number of steps equal to 
\bee
\overline{T}=\min_{T\in\mathbb{N}_+}\left\{T\left|\frac{\sum_{i\in\IB^{g_\citer}}A^{g_\citer}_i\gamma^{g_\citer}_{i,T}}{\sqrt{T}}\leq C\epsilon\right\}\right., 
\eee
with probability at least $1-\delta$, where $C$ is a constant independent of $T$. 
\label{thm:declare_inf}
\end{theorem}
\begin{proof}
Suppose up to step $T$, the optimization problem in line~\ref{alg_line:aux_prob} of Alg.~\ref{alg:constrained_grey_box_opt} keeps feasible. We have,
$$\bar{z}_{m^{g_k}}^{t,g_k}\leq0$$
and the corresponding real output $z^{t,g_k}_{m^{g_k}}$ corresponding to input $x_t$ satisfies 
$$z^{t,g_k}_{m^{g_k}}= g_\citer(x_t)\geq\epsilon$$
with probability at least $1-\delta$.

Accordingly, we have 
$$\sum_{t=1}^T\sum_{i\in\IB^{g_k}}A^{g_k}_i\beta^{g_k}_{i,t}\sigma_{i,t-1}({s}_{i})\geq\sum_{t=1}^T|g_\citer(x_t)-\bar{z}^{t,g_\citer}_{m^{g_\citer}}|\geq\sum_{t=1}^T\epsilon=T\epsilon\enspace,$$
where the first inequality follows by the output discrepancy bound in Lem.~\ref{lem:discrepancy_bound}. 
Meanwhile,
$$
\begin{aligned}
\sum_{t=1}^T\sum_{i\in\IB^{g_k}}A^{g_k}_i\beta^{g_k}_{i,t}\sigma_{i,t-1}({\bar{s}}_{i})&\leq\sum_{i\in\IB^{g_k}}A^{g_k}_i\beta^{g_k}_{i,T}\sum_{t=1}^T\sigma_{i,t-1}({\bar{s}}_{i})\\
&\leq \sum_{i\in\IB}A^{g_k}_i\beta^{g_k}_{i,T}\sqrt{4(T+2)\gamma^{g_k}_{i,T}}\enspace,
\end{aligned}
$$
where the last inequality follows by Lem.~\ref{lem:bound_cumu_sd}. Therefore,
$$\sum_{i\in\IB^{g_k}}A^{g_k}_i\beta^{g_k}_{i,T}\sqrt{4(T+2)\gamma^{g_k}_{i,T}}\geq T\epsilon,$$  
which implies $\epsilon\leq\frac{\sum_{i\in\IB^{g_k}}A^{g_k}_i\beta^{g_k}_{i,T}\sqrt{4(T+2)\gamma^{g_k}_{i,T}}}{T}$. 
By Def.~\ref{def:hpb_int}, we have 
$$\frac{\sum_{i\in\IB^{g_k}}A^{g_k}_i\beta^{g_k}_{i,T}\sqrt{4(T+2)\gamma^{g_k}_{i,T}}}{T}=\mathcal{O}\left(\frac{\sum_{i\in\IB^{g_\citer}}A^{g_\citer}_i\gamma^{g_\citer}_{i,T}}{\sqrt{T}}\right).$$
Hence, $\exists \tilde{C}>0$, such that,
\bee\epsilon\leq \tilde{C}\frac{\sum_{i\in\IB^{g_\citer}}A^{g_\citer}_i\gamma^{g_\citer}_{i,T}}{\sqrt{T}}.\eee
That is 
\bee C\epsilon\leq \frac{\sum_{i\in\IB^{g_\citer}}A^{g_\citer}_i\gamma^{g_\citer}_{i,T}}{\sqrt{T}},\label{inq:nece_cond_not_declare}\eee
where $C=\frac{1}{\tilde{C}}$.
However, since $\lim_{T\to\infty}\frac{\sum_{i\in\IB^{g_\citer}}A^{g_\citer}_i\gamma^{g_\citer}_{i,T}}{\sqrt{T}}=0$, the inequality~\eqref{inq:nece_cond_not_declare} will be violated when $T$ is large enough. So the optimistic auxiliary problem will be found infeasible on or before the first time the above inequality is violated, which is $\overline{T}$.      
 
\end{proof}

\begin{remark}
The proposed algorithm design and analysis in this paper focuses on the case where $T$ is given. In cases where $T$ is not given or not large enough to declare infeasibility, we can apply the `doubling trick'~\cite{besson2018doubling}. The basic idea is that we can start running a round of steps with $T=1$ and double $T$ after each round. As $T$ grows larger and larger, it will finally be larger than $\overline{T}$ and enough for infeasibility detection.  
\end{remark}

\section{Discussion on Special Cases}
\subsection{Grey-box function with only one layer of black-box functions}
We now restrict to the special case with only one layer of black-box functions. That is, the problem we consider can be formulated as, 
\begin{subequations}
\label{eqn:one_layer_prob_form}
\begin{eqnarray}
\min_{x\in\mathcal{X}}\quad &&F(x, \phi(x)),\label{eqn:one_l_comp_form}\\
\text{subject to}\quad && G_\citer(x, \phi(x))\leq 0\enspace,\quad \forall \citer\in[\tcon],\label{eqn:constraint}
\end{eqnarray}
\end{subequations}
where $\mathcal{X}\subset\mathbb{R}^n$ is a known compact subset of $\mathbb{R}^n$, $\phi:\mathbb{R}^n\to\mathbb{R}^{m-1}$, $F:\mathbb{R}^n\times\mathbb{R}^{m-1}\to\mathbb{R}$ and $G_k:\mathbb{R}^n\times\mathbb{R}^{m-1}\to\mathbb{R},~\citer\in[\tcon]$. We assume that the functions $F$ and $G_\citer,\citer\in[\tcon]$ are known and cheap to evaluate, and $\phi$ is a potentially vector-valued black box function that is expensive to evaluate. When restricted to this special case, our algorithm~\ref{alg:constrained_grey_box_opt} essentially recovers the algorithm proposed in~\cite{lu2023no}.

\section{Experiments}
\subsection{Linear Programming with Embedded Gaussian Process}
Linear programming~(LP) is a ubiquitous problem in engineering, operations research, and economics, etc. However, in many applications, LP problems may involve some unknown black-box functions other than the known linear part. In this part, we consider a generic linear programming problem with unknown expensive black-box functions embedded as shown in~\eqref{eqn:LP_GP}.
\begin{subequations}
\label{eqn:LP_GP}
\begin{eqnarray}
\min_{x\in\mathcal{X}}\quad &&c_1^Tx+c_2^Th(x),\label{eqn:LP_GP_obj}\\
\text{subject to}\quad && A_1x+A_2h(x)+b\leq 0\enspace.\label{eqn:LP_GP_constraint}
\end{eqnarray}
\end{subequations}
In this experiment, we consider $x\in[-2, 2]^2\subset\mathbb{R}^2$ and $h:\mathbb{R}^2\to\mathbb{R}^2$.

The corresponding auxiliary problem in the line~\ref{alg_line:aux_prob} of the Alg.~\ref{alg:constrained_grey_box_opt} is reduced to, 
\begin{subequations}
\label{eqn:LP_GP_aux}
\begin{eqnarray}
\min_{x\in\mathcal{X}}\min_{z\in[l_t(x), u_t(x)]}\quad &&c_1^Tx+c_2^Tz,\label{eqn:LP_GP_aux_obj}\\
\text{subject to}\quad && A_1x+A_2z+b\leq 0\enspace.\label{eqn:LP_GP_aux_constraint_1}
\end{eqnarray}
\end{subequations}
To compare the performance of different algorithms in terms of the speed of finding the global optimum, we also introduce the metric of constrained regret defined as, 
\begin{equation}
    \mathrm{CR}_t\coloneqq\min_{\tau\in[t]}[f(x_\tau)-f(x^*)]^++\|[g(x_\tau)]^+\|_1,
\end{equation}
where $x^*$ is the ground truth optimal solution.
To evaluate the performance of different algorithms, we randomly sample $c_1,c_2,b$ and the rows of $A_1, A_2$ uniformly from the sphere $\mathbb{S}^1$. And we sample the function $h$ from a Gaussian process with kernel function,
\begin{equation}
k(x,y)=\frac{1}{2}\exp\{-(x_1-y_1)^2-(x_2-y_2)^2\}. 
\end{equation}

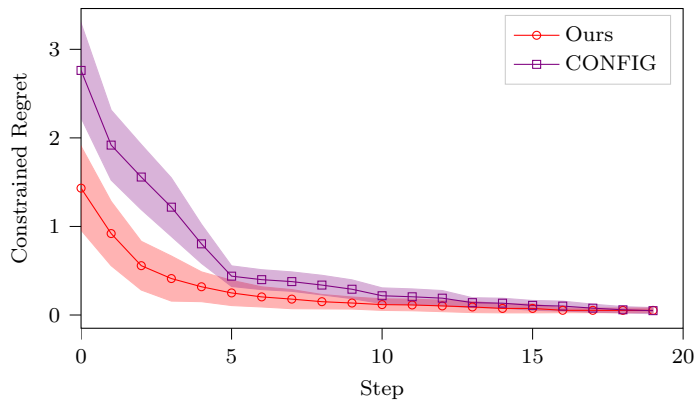
\begin{figure}[htbp!]
    \centering
\begin{tikzpicture}

\definecolor{darkgray176}{RGB}{176,176,176}
\definecolor{lightgray204}{RGB}{204,204,204}
\definecolor{purple}{RGB}{128,0,128}

\begin{axis}[
legend cell align={left},
legend style={fill opacity=0.8, draw opacity=1, text opacity=1, draw=lightgray204},
tick align=outside,
tick pos=left,
x grid style={darkgray176},
xlabel={Step},
xmin=0, xmax=20,
xtick style={color=black},
xtick={0,5,10,15,20},
xticklabels={
  \(\displaystyle {0}\),
  \(\displaystyle {5}\),
  \(\displaystyle {10}\),
  \(\displaystyle {15}\),
  \(\displaystyle {20}\)
},
y grid style={darkgray176},
ylabel={Constrained Regret},
ymin=-0.149585684284211, ymax=3.46077903577686,
ytick style={color=black},
ytick={-1,0,1,2,3,4},
yticklabels={
  \(\displaystyle {\ensuremath{-}1}\),
  \(\displaystyle {0}\),
  \(\displaystyle {1}\),
  \(\displaystyle {2}\),
  \(\displaystyle {3}\),
  \(\displaystyle {4}\)
}
]
\path [draw=red, fill=red, opacity=0.3]
(axis cs:0,1.90594594119865)
--(axis cs:0,0.958550335210325)
--(axis cs:1,0.556906844436765)
--(axis cs:2,0.281976174791794)
--(axis cs:3,0.157165337124393)
--(axis cs:4,0.150668237641646)
--(axis cs:5,0.106730043162484)
--(axis cs:6,0.0907193090587717)
--(axis cs:7,0.0712665016084746)
--(axis cs:8,0.0691056590429524)
--(axis cs:9,0.0673507094740478)
--(axis cs:10,0.0520572724215219)
--(axis cs:11,0.0479375009929485)
--(axis cs:12,0.0379266083153201)
--(axis cs:13,0.0272991409031766)
--(axis cs:14,0.0241805717311188)
--(axis cs:15,0.0241805717311188)
--(axis cs:16,0.0264116786730708)
--(axis cs:17,0.0262884230791116)
--(axis cs:18,0.0260375635430127)
--(axis cs:19,0.0260375635430127)
--(axis cs:19,0.0784966539780193)
--(axis cs:19,0.0784966539780193)
--(axis cs:18,0.0784966539780193)
--(axis cs:17,0.0793983143565303)
--(axis cs:16,0.0798954609541822)
--(axis cs:15,0.124721097246489)
--(axis cs:14,0.124721097246489)
--(axis cs:13,0.154122516741266)
--(axis cs:12,0.16748281730629)
--(axis cs:11,0.178955054809517)
--(axis cs:10,0.182790836334284)
--(axis cs:9,0.203370875572274)
--(axis cs:8,0.231906275777843)
--(axis cs:7,0.28726720101038)
--(axis cs:6,0.318402754117391)
--(axis cs:5,0.391895355907057)
--(axis cs:4,0.487234590917907)
--(axis cs:3,0.666184811849939)
--(axis cs:2,0.832135592842961)
--(axis cs:1,1.28246081345309)
--(axis cs:0,1.90594594119865)
--cycle;

\path [draw=purple, fill=purple, opacity=0.3]
(axis cs:0,3.29667154850135)
--(axis cs:0,2.22723590813417)
--(axis cs:1,1.5234602609425)
--(axis cs:2,1.19034698658497)
--(axis cs:3,0.887092063697166)
--(axis cs:4,0.582981840688975)
--(axis cs:5,0.322158563533558)
--(axis cs:6,0.286804615407698)
--(axis cs:7,0.267642548317288)
--(axis cs:8,0.227395865054255)
--(axis cs:9,0.184910322869573)
--(axis cs:10,0.128515986085305)
--(axis cs:11,0.117046410209796)
--(axis cs:12,0.104004109003601)
--(axis cs:13,0.0837174197843795)
--(axis cs:14,0.0774461420508302)
--(axis cs:15,0.0536163277562081)
--(axis cs:16,0.0448539113730877)
--(axis cs:17,0.0323052626528768)
--(axis cs:18,0.0228340801043097)
--(axis cs:19,0.014521802991292)
--(axis cs:19,0.0835629398981377)
--(axis cs:19,0.0835629398981377)
--(axis cs:18,0.0952362684640249)
--(axis cs:17,0.122458883307538)
--(axis cs:16,0.156811516812139)
--(axis cs:15,0.164381318977839)
--(axis cs:14,0.189507827403383)
--(axis cs:13,0.195570954790793)
--(axis cs:12,0.276258224282961)
--(axis cs:11,0.294768976075838)
--(axis cs:10,0.306503723594333)
--(axis cs:9,0.397091617939454)
--(axis cs:8,0.448510250660836)
--(axis cs:7,0.486917721946308)
--(axis cs:6,0.51168872855252)
--(axis cs:5,0.555151468129644)
--(axis cs:4,1.02412739177712)
--(axis cs:3,1.54826327663312)
--(axis cs:2,1.92542620948783)
--(axis cs:1,2.31365334849806)
--(axis cs:0,3.29667154850135)
--cycle;

\addplot [red, mark=o, mark size=\markSize, mark options={solid,fill=none}]
table {%
0 1.43224811553955
1 0.919683814048767
2 0.557055950164795
3 0.411675095558167
4 0.318951368331909
5 0.24931263923645
6 0.204560995101929
7 0.179266810417175
8 0.150506019592285
9 0.135360836982727
10 0.117424011230469
11 0.113446235656738
12 0.102704763412476
13 0.0907108783721924
14 0.0744508504867554
15 0.0744508504867554
16 0.0531535148620605
17 0.0528433322906494
18 0.0522670745849609
19 0.0522670745849609
};
\addlegendentry{Ours}
\addplot [purple, mark=square, mark size=\markSize, mark options={solid,fill=none}]
table {%
0 2.76195383071899
1 1.91855680942535
2 1.55788660049438
3 1.21767771244049
4 0.803554534912109
5 0.438655018806458
6 0.399246692657471
7 0.377280116081238
8 0.337953090667725
9 0.291000962257385
10 0.217509865760803
11 0.205907702445984
12 0.190131187438965
13 0.139644145965576
14 0.133476972579956
15 0.108998775482178
16 0.10083270072937
17 0.0773820877075195
18 0.0590351819992065
19 0.0490423440933228
};
\addlegendentry{CONFIG}
\end{axis}

\end{tikzpicture}
    \caption{The evolution of constrained regret with respect to sample step.}
    \label{fig:constr_regret}
\end{figure}

We compare our grey-box optimization method to the constrained black-box optimization method~\cite{xu2022constrained}. Fig.~\ref{fig:constr_regret} to Fig.~\ref{fig:cumu_vio_2} show the evolutions of constrained regret, cumulative positive regret, cumulative violations for constraint 1 and constraint 2. It can be observed that our grey-box optimization method significantly outperforms the pure black-box optimization method in terms of all the above metrics.   
\begin{figure}[htbp!]
    \centering
\begin{tikzpicture}

\definecolor{darkgray176}{RGB}{176,176,176}
\definecolor{lightgray204}{RGB}{204,204,204}
\definecolor{purple}{RGB}{128,0,128}

\begin{axis}[
legend cell align={left},
legend style={
  fill opacity=0.8,
  draw opacity=1,
  text opacity=1,
  at={(0.03,0.97)},
  anchor=north west,
  draw=lightgray204
},
tick align=outside,
tick pos=left,
x grid style={darkgray176},
xlabel={Step},
xmin=0, xmax=21,
xtick style={color=black},
xtick={0,5,10,15,20,25},
xticklabels={
  \(\displaystyle {0}\),
  \(\displaystyle {5}\),
  \(\displaystyle {10}\),
  \(\displaystyle {15}\),
  \(\displaystyle {20}\),
  \(\displaystyle {25}\)
},
y grid style={darkgray176},
ylabel={Cumulative Positive Regret},
ymin=-0.556533169244694, ymax=11.6871965541386,
ytick style={color=black},
ytick={-2.5,0,2.5,5,7.5,10,12.5},
yticklabels={
  \(\displaystyle {\ensuremath{-}2.5}\),
  \(\displaystyle {0.0}\),
  \(\displaystyle {2.5}\),
  \(\displaystyle {5.0}\),
  \(\displaystyle {7.5}\),
  \(\displaystyle {10.0}\),
  \(\displaystyle {12.5}\)
}
]
\path [draw=red, fill=red, opacity=0.3]
(axis cs:0,0)
--(axis cs:0,0)
--(axis cs:1,0.041697095218929)
--(axis cs:2,0.298944066823616)
--(axis cs:3,0.5502970875516)
--(axis cs:4,0.794431601434799)
--(axis cs:5,1.10442302698425)
--(axis cs:6,1.47499153759809)
--(axis cs:7,1.69765815054166)
--(axis cs:8,1.9437182027864)
--(axis cs:9,2.20345108742546)
--(axis cs:10,2.41110540295179)
--(axis cs:11,2.47900166404148)
--(axis cs:12,2.67638862449953)
--(axis cs:13,2.79114107292223)
--(axis cs:14,2.84729924256916)
--(axis cs:15,2.85326043853524)
--(axis cs:16,2.92800193034626)
--(axis cs:17,2.93791420830508)
--(axis cs:18,2.93864005477394)
--(axis cs:19,2.98917317873673)
--(axis cs:20,2.99281465518336)
--(axis cs:20,4.90619980634888)
--(axis cs:20,4.90619980634888)
--(axis cs:19,4.90192459678855)
--(axis cs:18,4.84946901709424)
--(axis cs:17,4.84868619929354)
--(axis cs:16,4.83981726808975)
--(axis cs:15,4.77361914394015)
--(axis cs:14,4.76949130779393)
--(axis cs:13,4.7008075830666)
--(axis cs:12,4.59364627409682)
--(axis cs:11,4.22469782608337)
--(axis cs:10,4.15619952240438)
--(axis cs:9,3.91989738099999)
--(axis cs:8,3.51295217832753)
--(axis cs:7,3.16432969638606)
--(axis cs:6,2.74243194944895)
--(axis cs:5,2.07685420900889)
--(axis cs:4,1.52848904299578)
--(axis cs:3,1.09312899799206)
--(axis cs:2,0.677441376515541)
--(axis cs:1,0.192340820942568)
--(axis cs:0,0)
--cycle;

\path [draw=purple, fill=purple, opacity=0.3]
(axis cs:0,0)
--(axis cs:0,0)
--(axis cs:1,1.26739810722875)
--(axis cs:2,2.07006515472124)
--(axis cs:3,2.65247497247831)
--(axis cs:4,2.92683858246488)
--(axis cs:5,3.10273930290717)
--(axis cs:6,3.21331909510261)
--(axis cs:7,3.67087432874002)
--(axis cs:8,4.3441562150296)
--(axis cs:9,5.17401566682465)
--(axis cs:10,5.72382538950141)
--(axis cs:11,6.0313433277535)
--(axis cs:12,6.48863340088384)
--(axis cs:13,6.92622040725204)
--(axis cs:14,7.54606087479525)
--(axis cs:15,7.91491441712325)
--(axis cs:16,8.05335890684742)
--(axis cs:17,8.21443238323727)
--(axis cs:18,8.24541619257135)
--(axis cs:19,8.27895166511295)
--(axis cs:20,8.38659317246169)
--(axis cs:20,11.1306633848939)
--(axis cs:20,11.1306633848939)
--(axis cs:19,11.0607363150626)
--(axis cs:18,11.0201721883275)
--(axis cs:17,10.9983757829638)
--(axis cs:16,10.8444894368874)
--(axis cs:15,10.6815529159345)
--(axis cs:14,10.246966056708)
--(axis cs:13,9.46727882986939)
--(axis cs:12,8.95695183255747)
--(axis cs:11,8.52344832059351)
--(axis cs:10,8.18143134948774)
--(axis cs:9,7.61539832444371)
--(axis cs:8,6.66596834152085)
--(axis cs:7,5.96916238441485)
--(axis cs:6,5.46158762505131)
--(axis cs:5,5.27803941820919)
--(axis cs:4,4.99866457331927)
--(axis cs:3,4.5333389205141)
--(axis cs:2,3.32785233351611)
--(axis cs:1,2.23982566584519)
--(axis cs:0,0)
--cycle;

\addplot [red, mark=o, mark size=\markSize, mark options={solid,fill=none}]
table {%
0 0
1 0.117018938064575
2 0.488192677497864
3 0.821712970733643
4 1.1614602804184
5 1.59063863754272
6 2.10871171951294
7 2.43099403381348
8 2.72833514213562
9 3.06167411804199
10 3.28365254402161
11 3.35184979438782
12 3.63501739501953
13 3.74597430229187
14 3.80839538574219
15 3.81343984603882
16 3.88390970230103
17 3.8933002948761
18 3.89405465126038
19 3.94554877281189
20 3.94950723648071
};
\addlegendentry{Ours}
\addplot [purple, mark=square, mark size=\markSize, mark options={solid,fill=none}]
table {%
0 0
1 1.7536119222641
2 2.6989586353302
3 3.5929069519043
4 3.96275162696838
5 4.19038915634155
6 4.33745336532593
7 4.82001829147339
8 5.50506210327148
9 6.39470720291138
10 6.95262813568115
11 7.27739572525024
12 7.72279262542725
13 8.19674968719482
14 8.89651393890381
15 9.29823398590088
16 9.44892406463623
17 9.60640430450439
18 9.63279438018799
19 9.66984367370605
20 9.75862789154053
};
\addlegendentry{CONFIG}
\end{axis}

\end{tikzpicture}
    \caption{The evolution of cumulative positive regret with respect to sample step.}
    \label{fig:cumu_pos_r}
\end{figure}
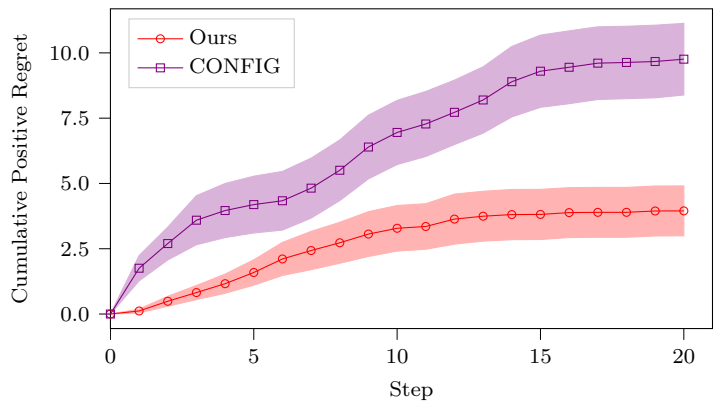
\begin{figure}[htbp!]
    \centering
\begin{tikzpicture}

\definecolor{darkgray176}{RGB}{176,176,176}
\definecolor{lightgray204}{RGB}{204,204,204}
\definecolor{purple}{RGB}{128,0,128}

\begin{axis}[
legend cell align={left},
legend style={
  fill opacity=0.8,
  draw opacity=1,
  text opacity=1,
  at={(0.03,0.97)},
  anchor=north west,
  draw=lightgray204
},
tick align=outside,
tick pos=left,
x grid style={darkgray176},
xlabel={Step},
xmin=0, xmax=21,
xtick style={color=black},
xtick={0,5,10,15,20,25},
xticklabels={
  \(\displaystyle {0}\),
  \(\displaystyle {5}\),
  \(\displaystyle {10}\),
  \(\displaystyle {15}\),
  \(\displaystyle {20}\),
  \(\displaystyle {25}\)
},
y grid style={darkgray176},
ylabel={Cumulative Constraint Violation 1},
ymin=-0.472191854359757, ymax=9.9160289415549,
ytick style={color=black},
ytick={-2.5,0,2.5,5,7.5,10},
yticklabels={
  \(\displaystyle {\ensuremath{-}2.5}\),
  \(\displaystyle {0.0}\),
  \(\displaystyle {2.5}\),
  \(\displaystyle {5.0}\),
  \(\displaystyle {7.5}\),
  \(\displaystyle {10.0}\)
}
]
\path [draw=red, fill=red, opacity=0.3]
(axis cs:0,0)
--(axis cs:0,0)
--(axis cs:1,0.181030554585136)
--(axis cs:2,0.476914723661356)
--(axis cs:3,0.822806247667602)
--(axis cs:4,0.955177358867345)
--(axis cs:5,1.08607424187555)
--(axis cs:6,1.35755144380341)
--(axis cs:7,1.50180759411618)
--(axis cs:8,1.59774224660195)
--(axis cs:9,1.74760039573367)
--(axis cs:10,1.84595735557326)
--(axis cs:11,1.92704522344475)
--(axis cs:12,2.02067215070929)
--(axis cs:13,2.10783553465785)
--(axis cs:14,2.20331686313774)
--(axis cs:15,2.2888784822055)
--(axis cs:16,2.37316979514401)
--(axis cs:17,2.41988196416318)
--(axis cs:18,2.44835420634869)
--(axis cs:19,2.51585085135009)
--(axis cs:20,2.54217575359883)
--(axis cs:20,4.95781178189827)
--(axis cs:20,4.95781178189827)
--(axis cs:19,4.89427872986744)
--(axis cs:18,4.75912892799942)
--(axis cs:17,4.70052970704084)
--(axis cs:16,4.60483582011271)
--(axis cs:15,4.49107839980053)
--(axis cs:14,4.35754011296711)
--(axis cs:13,4.26784109132447)
--(axis cs:12,4.10453145084024)
--(axis cs:11,3.95273571772038)
--(axis cs:10,3.87680085931871)
--(axis cs:9,3.76101407687802)
--(axis cs:8,3.60589725127363)
--(axis cs:7,3.38967931635945)
--(axis cs:6,3.03136989242127)
--(axis cs:5,2.6294605131774)
--(axis cs:4,2.30362119180812)
--(axis cs:3,1.96206720166577)
--(axis cs:2,1.28414884011765)
--(axis cs:1,0.618278677451719)
--(axis cs:0,0)
--cycle;

\path [draw=purple, fill=purple, opacity=0.3]
(axis cs:0,0)
--(axis cs:0,0)
--(axis cs:1,0.142609432417726)
--(axis cs:2,0.824546022324205)
--(axis cs:3,1.51769066771589)
--(axis cs:4,1.97378796058102)
--(axis cs:5,2.44409300222525)
--(axis cs:6,2.76534457488152)
--(axis cs:7,3.19735877037879)
--(axis cs:8,3.38856742337703)
--(axis cs:9,3.64079541986412)
--(axis cs:10,4.07123956807169)
--(axis cs:11,4.35403988080609)
--(axis cs:12,4.7565458965103)
--(axis cs:13,4.93755168772781)
--(axis cs:14,5.01404882738652)
--(axis cs:15,5.31179789404698)
--(axis cs:16,5.52859418252655)
--(axis cs:17,5.72249869812842)
--(axis cs:18,5.77927420901085)
--(axis cs:19,5.87676697917728)
--(axis cs:20,5.95072220468918)
--(axis cs:20,9.44383708719514)
--(axis cs:20,9.44383708719514)
--(axis cs:19,9.32039838459333)
--(axis cs:18,9.13777820571361)
--(axis cs:17,9.04716012476824)
--(axis cs:16,8.71061813238865)
--(axis cs:15,8.343039376305)
--(axis cs:14,8.05448949440886)
--(axis cs:13,7.94714245560557)
--(axis cs:12,7.68005338230743)
--(axis cs:11,7.19197338185826)
--(axis cs:10,6.68154532052234)
--(axis cs:9,6.18597831932649)
--(axis cs:8,5.82343630244804)
--(axis cs:7,5.49662561591709)
--(axis cs:6,4.86008580962133)
--(axis cs:5,4.29991018374802)
--(axis cs:4,3.59643853677033)
--(axis cs:3,2.72546938934659)
--(axis cs:2,1.70503852388469)
--(axis cs:1,0.619128985200123)
--(axis cs:0,0)
--cycle;

\addplot [red, mark=o, mark size=\markSize, mark options={solid,fill=none}]
table {%
0 0
1 0.399654626846313
2 0.880531787872314
3 1.39243674278259
4 1.62939929962158
5 1.85776734352112
6 2.19446063041687
7 2.44574356079102
8 2.60181975364685
9 2.75430727005005
10 2.86137914657593
11 2.93989038467407
12 3.06260180473328
13 3.18783831596375
14 3.28042840957642
15 3.38997840881348
16 3.48900270462036
17 3.56020593643188
18 3.60374164581299
19 3.70506477355957
20 3.74999380111694
};
\addlegendentry{Ours}
\addplot [purple, mark=square, mark size=\markSize, mark options={solid,fill=none}]
table {%
0 0
1 0.380869150161743
2 1.26479232311249
3 2.12158012390137
4 2.78511333465576
5 3.37200164794922
6 3.81271529197693
7 4.34699201583862
8 4.60600185394287
9 4.91338682174683
10 5.37639236450195
11 5.77300643920898
12 6.21829986572266
13 6.44234704971313
14 6.53426933288574
15 6.8274188041687
16 7.11960601806641
17 7.3848295211792
18 7.45852613449097
19 7.59858274459839
20 7.69727945327759
};
\addlegendentry{CONFIG}
\end{axis}

\end{tikzpicture}
    \caption{The evolution of the cumulative violation for the first constraint with respect to sample step.}
    \label{fig:cumu_vio_1}
\end{figure}
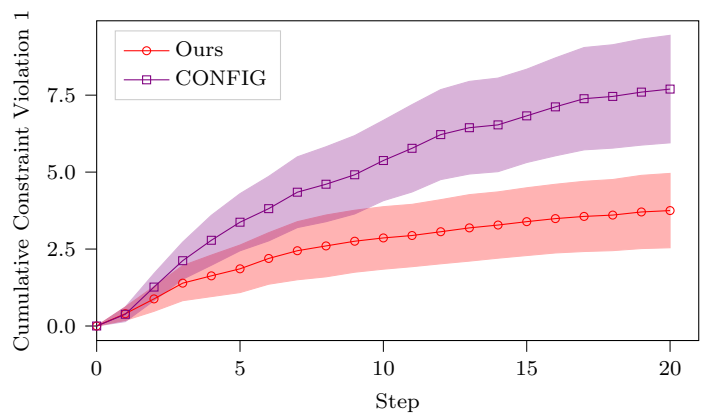
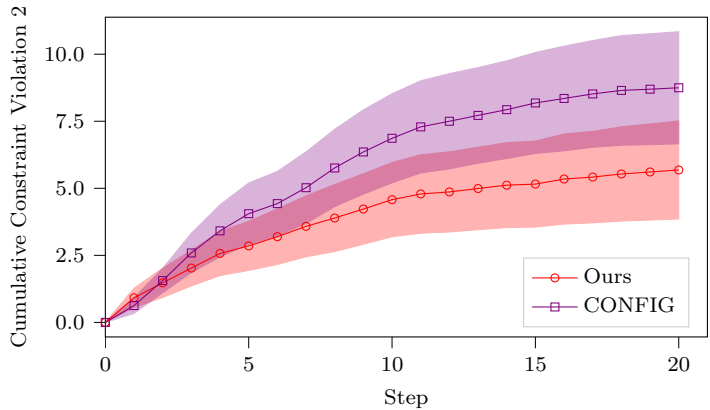
\begin{figure}[htbp!]
    \centering
\begin{tikzpicture}

\definecolor{darkgray176}{RGB}{176,176,176}
\definecolor{lightgray204}{RGB}{204,204,204}
\definecolor{purple}{RGB}{128,0,128}

\begin{axis}[
legend cell align={left},
legend style={
  fill opacity=0.8,
  draw opacity=1,
  text opacity=1,
  at={(0.97,0.03)},
  anchor=south east,
  draw=lightgray204
},
tick align=outside,
tick pos=left,
x grid style={darkgray176},
xlabel={Step},
xmin=0, xmax=21,
xtick style={color=black},
xtick={0,5,10,15,20,25},
xticklabels={
  \(\displaystyle {0}\),
  \(\displaystyle {5}\),
  \(\displaystyle {10}\),
  \(\displaystyle {15}\),
  \(\displaystyle {20}\),
  \(\displaystyle {25}\)
},
y grid style={darkgray176},
ylabel={Cumulative Constraint Violation 2},
ymin=-0.541888361492078, ymax=11.3796555913336,
ytick style={color=black},
ytick={-2.5,0,2.5,5,7.5,10,12.5},
yticklabels={
  \(\displaystyle {\ensuremath{-}2.5}\),
  \(\displaystyle {0.0}\),
  \(\displaystyle {2.5}\),
  \(\displaystyle {5.0}\),
  \(\displaystyle {7.5}\),
  \(\displaystyle {10.0}\),
  \(\displaystyle {12.5}\)
}
]
\path [draw=red, fill=red, opacity=0.3]
(axis cs:0,0)
--(axis cs:0,0)
--(axis cs:1,0.557297406948862)
--(axis cs:2,0.929514915187469)
--(axis cs:3,1.36072114200354)
--(axis cs:4,1.74658320401441)
--(axis cs:5,1.93699485562941)
--(axis cs:6,2.16063292235711)
--(axis cs:7,2.44473364068448)
--(axis cs:8,2.6411213487694)
--(axis cs:9,2.91329190991988)
--(axis cs:10,3.19405046497462)
--(axis cs:11,3.32627773477963)
--(axis cs:12,3.37193148070999)
--(axis cs:13,3.45598814243218)
--(axis cs:14,3.53585020753121)
--(axis cs:15,3.55325781141486)
--(axis cs:16,3.66480075530953)
--(axis cs:17,3.7138174970388)
--(axis cs:18,3.77757790463528)
--(axis cs:19,3.81780809052146)
--(axis cs:20,3.85562184454566)
--(axis cs:20,7.51147629726103)
--(axis cs:20,7.51147629726103)
--(axis cs:19,7.39946633452737)
--(axis cs:18,7.29000788967994)
--(axis cs:17,7.12361816931641)
--(axis cs:16,7.02595536548323)
--(axis cs:15,6.76033401375918)
--(axis cs:14,6.6978168987862)
--(axis cs:13,6.5289579490824)
--(axis cs:12,6.35760872933083)
--(axis cs:11,6.25268725942486)
--(axis cs:10,5.95850011949886)
--(axis cs:9,5.54055353041528)
--(axis cs:8,5.14141863830666)
--(axis cs:7,4.71758927157268)
--(axis cs:6,4.23309605269533)
--(axis cs:5,3.7659422825249)
--(axis cs:4,3.39608542900353)
--(axis cs:3,2.68930971511799)
--(axis cs:2,2.02211476063313)
--(axis cs:1,1.27385172126176)
--(axis cs:0,0)
--cycle;

\path [draw=purple, fill=purple, opacity=0.3]
(axis cs:0,0)
--(axis cs:0,0)
--(axis cs:1,0.336662138526022)
--(axis cs:2,1.1056853709598)
--(axis cs:3,1.8571445005732)
--(axis cs:4,2.44326265964226)
--(axis cs:5,2.91237052266336)
--(axis cs:6,3.22984817866454)
--(axis cs:7,3.68761854340856)
--(axis cs:8,4.31890650777237)
--(axis cs:9,4.78557811851251)
--(axis cs:10,5.2033884251296)
--(axis cs:11,5.57353053576935)
--(axis cs:12,5.72429841635022)
--(axis cs:13,5.9350035035286)
--(axis cs:14,6.11154349819003)
--(axis cs:15,6.29983396652293)
--(axis cs:16,6.39943044416116)
--(axis cs:17,6.52977691118674)
--(axis cs:18,6.60594734524864)
--(axis cs:19,6.62867980478754)
--(axis cs:20,6.65852203189409)
--(axis cs:20,10.8377672298416)
--(axis cs:20,10.8377672298416)
--(axis cs:19,10.7617286258759)
--(axis cs:18,10.6926042394336)
--(axis cs:17,10.5085348950684)
--(axis cs:16,10.2995148365951)
--(axis cs:15,10.0631260433622)
--(axis cs:14,9.75360139404385)
--(axis cs:13,9.50073111814086)
--(axis cs:12,9.27504834882153)
--(axis cs:11,9.0078179323615)
--(axis cs:10,8.52495357549522)
--(axis cs:9,7.92511006045587)
--(axis cs:8,7.19783958269813)
--(axis cs:7,6.35738806449011)
--(axis cs:6,5.62861824973456)
--(axis cs:5,5.19561342379693)
--(axis cs:4,4.38521386365104)
--(axis cs:3,3.32260198639627)
--(axis cs:2,2.01371812223813)
--(axis cs:1,0.918283127417711)
--(axis cs:0,0)
--cycle;

\addplot [red, mark=o, mark size=\markSize, mark options={solid,fill=none}]
table {%
0 0
1 0.915574550628662
2 1.47581481933594
3 2.02501535415649
4 2.57133436203003
5 2.85146856307983
6 3.19686460494995
7 3.58116149902344
8 3.89126992225647
9 4.22692251205444
10 4.57627534866333
11 4.78948259353638
12 4.86476993560791
13 4.99247312545776
14 5.11683368682861
15 5.15679597854614
16 5.34537792205811
17 5.41871786117554
18 5.5337929725647
19 5.60863733291626
20 5.68354892730713
};
\addlegendentry{Ours}
\addplot [purple, mark=square, mark size=\markSize, mark options={solid,fill=none}]
table {%
0 0
1 0.627472639083862
2 1.55970180034637
3 2.58987331390381
4 3.4142382144928
5 4.05399179458618
6 4.42923307418823
7 5.02250337600708
8 5.75837326049805
9 6.35534429550171
10 6.86417102813721
11 7.29067420959473
12 7.49967336654663
13 7.71786737442017
14 7.93257236480713
15 8.18148040771484
16 8.34947299957275
17 8.51915550231934
18 8.64927577972412
19 8.69520378112793
20 8.74814510345459
};
\addlegendentry{CONFIG}
\end{axis}

\end{tikzpicture}
    \caption{The evolution of cumulative violation for the second constraint with respect to sample step.}
    \label{fig:cumu_vio_2}
\end{figure}

\section{Conclusion}
In this paper, we have formulated a general \emph{grey-box} optimization problem, i.e., with both objective and constraint functions composed of both \emph{black-box} and \emph{white-box} functions. Our formulation covers the existing grey-box optimization formulations as special cases. An \emph{optimism}-driven algorithm is proposed to efficiently solve it. Under certain regularity assumptions, our algorithm achieves similar regret bound as that for the standard black-box Bayesian optimization algorithm, up to a constant term depending on the Lipschitz constants of the functions considered. The method is then extended to the constrained case and special cases are discussed. For the commonly used kernel functions, kernel-specific convergence rate to the optimal solution is derived. Experimental results show that our grey-box optimization method empirically improves the speed of finding the global optimum solution significantly, as compared to the standard black-box optimization algorithm.    


\bibliographystyle{plain} 
\bibliography{refs.bib}

\end{document}